\documentclass[final]{cvpr}
\PassOptionsToPackage{numbers, compress}{natbib}

\usepackage{times}
\usepackage{epsfig}
\usepackage{graphicx}
\usepackage{amsmath}
\usepackage{amssymb}

\usepackage{caption}    
\usepackage{subcaption} 
\usepackage[usestackEOL]{stackengine}
\usepackage{wrapfig}    
\usepackage{diagbox}    
\usepackage{multirow}
\usepackage{booktabs}
\usepackage{bm}
\usepackage[table]{xcolor}
\definecolor{mygray}{gray}{0.9}
\usepackage[pagebackref=true,breaklinks=true,colorlinks,bookmarks=false]{hyperref}

\def\cvprPaperID{4951} 
\def\confYear{CVPR 2021}

\newcommand{\norm}[1]{\left\lVert#1\right\rVert}

\author{%
 \textbf{Zhongqi Yue}\textsuperscript{1,3}\thanks{Equal contribution}, \quad \textbf{Tan Wang}\textsuperscript{1}\footnotemark[1], \quad \textbf{Hanwang Zhang}\textsuperscript{1}, \quad \textbf{Qianru Sun}\textsuperscript{2}, \quad \textbf{Xian-Sheng Hua}\textsuperscript{3}\\
\small \textsuperscript{1}Nanyang Technological University,\quad \textsuperscript{2}Singapore Management University, \small\textsuperscript{3}Damo Academy, Alibaba Group\\
\tt\small yuez0003@ntu.edu.sg,\quad TAN317@e.ntu.edu.sg,\quad hanwangzhang@ntu.edu.sg,\\
\tt\small \quad qianrusun@smu.edu.sg, \quad xiansheng.hxs@alibaba-inc.com\\}

\begin{document}

\title{Counterfactual Zero-Shot and Open-Set Visual Recognition}



\maketitle

\begin{abstract}
We present a novel counterfactual framework for both Zero-Shot Learning (ZSL) and Open-Set Recognition (OSR), whose common challenge is generalizing to the unseen-classes by only training on the seen-classes. Our idea stems from the observation that the generated samples for unseen-classes are often out of the true distribution, which causes severe recognition rate imbalance between the seen-class (high) and unseen-class (low). We show that the key reason is that the generation is not \textbf{Counterfactual Faithful}, and thus we propose a faithful one, whose generation is from the sample-specific counterfactual question: What would the sample look like, if we set its class attribute to a certain class, while keeping its sample attribute unchanged? Thanks to the faithfulness, we can apply the \textbf{Consistency Rule} to perform unseen/seen binary classification, by asking: Would its counterfactual still look like itself? If ``yes'', the sample is from a certain class, and ``no'' otherwise. Through extensive experiments on ZSL and OSR, we demonstrate that our framework effectively mitigates the seen/unseen imbalance and hence significantly improves the overall performance. Note that this framework is orthogonal to existing methods, thus, it can serve as a new baseline to evaluate how ZSL/OSR models generalize. Codes are available at \url{https://github.com/yue-zhongqi/gcm-cf}.
\end{abstract}
\vspace{-5mm}
\section{Introduction}
\label{sec:1}

Generalizing visual recognition to novel classes unseen in training is perhaps the Holy Grail of machine vision~\cite{marr2010vision}. For example, if machines could classify new classes accurately by Zero-Shot Learning (ZSL\footnote[1]{Conventional ZSL~\cite{lampert2009learning} only evaluates the recognition on the unseen-classes. We refer to ZSL as a more challenging setting: Generalized ZSL~\cite{chao2016empirical}, which is evaluated on the both seen- and unseen-classes.})~\cite{lampert2009learning, xian2018zero}, we could collect labelled data as many as possible for free; if machines could reject samples of unknown classes by Open-Set Recognition (OSR)~\cite{scheirer2014probability,cardoso2015bounded}, any recognition system would be shielded against outliers. Unfortunately, this goal is far from achieved, as it is yet a great challenge for them to ``imagine'' the unseen world based on the seen one~\cite{lake2017building,scholkopf2019causality}.
\begin{figure}
    \centering
    \footnotesize
    \includegraphics[width=.47\textwidth]{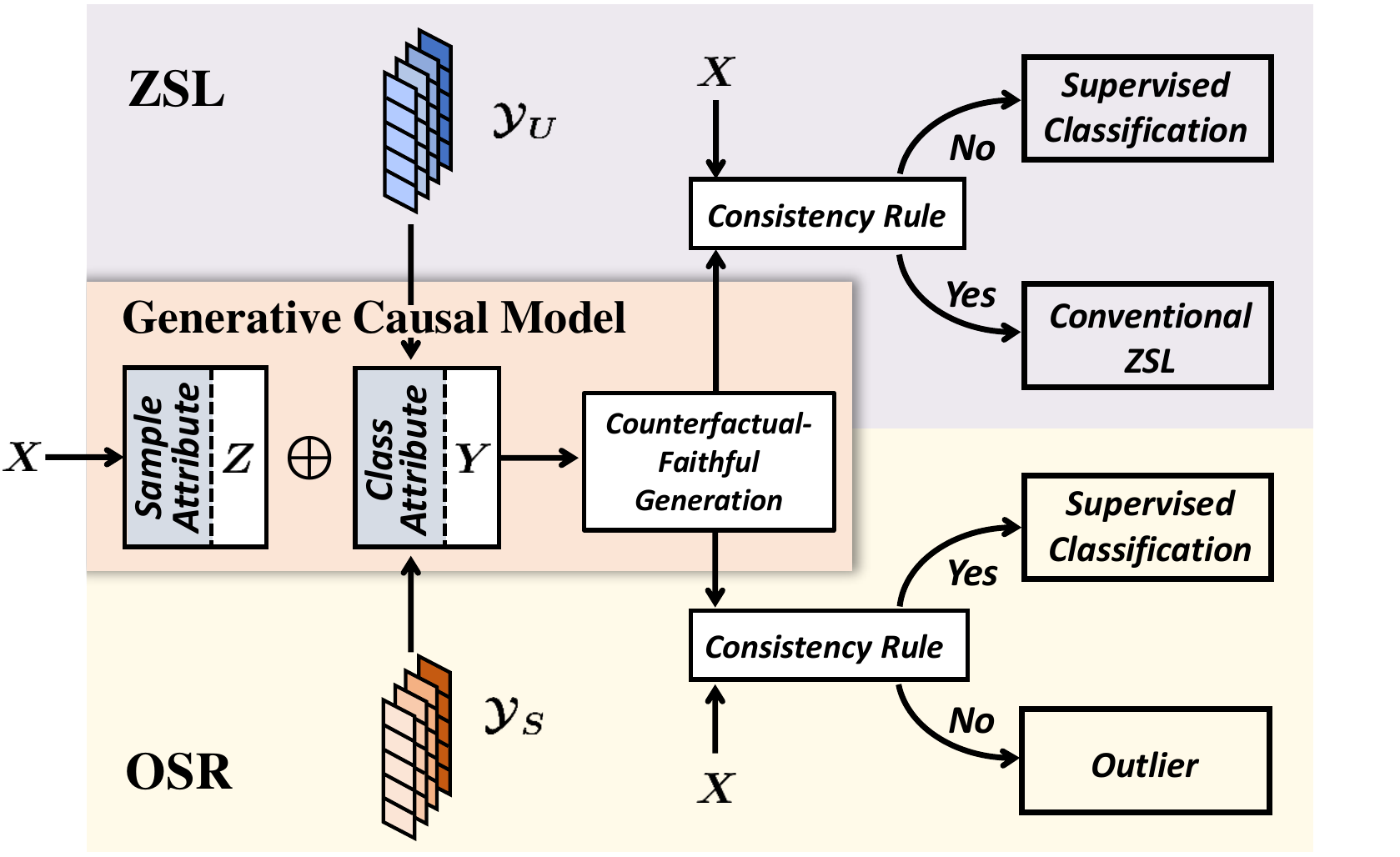}
    \captionof{figure}{Our counterfactual framework for ZSL and OSR. Here $\oplus$ denotes vector concatenation, $\mathcal{Y}_U$ and $\mathcal{Y}_S$ denote the set of unseen and seen class attributes, respectively.}
    \vspace{-6mm}
    \label{fig:1}
\end{figure}

Over the past decade, all the unseen-class recognition methods stem from the same grand assumption: \emph{attributes (or features) learned from the training seen-classes are transferable to the testing unseen-classes}. Therefore, if we have the ground-truth \emph{class attributes} describing both of the seen- and unseen-classes (or only those of the seen in OSR), ZSL (or OSR) can be accomplished by comparing the predicted class attributes of the test sample and the ground-truth ones~\cite{frome2013devise, bendale2016towards}; or by training a classifier on the samples generated from the ground-truth attributes~\cite{xian2019f, oza2019c2ae}.

\begin{figure*}
    \centering
    \footnotesize
    \begin{subfigure}[t]{\textwidth}
         \includegraphics[width=\textwidth]{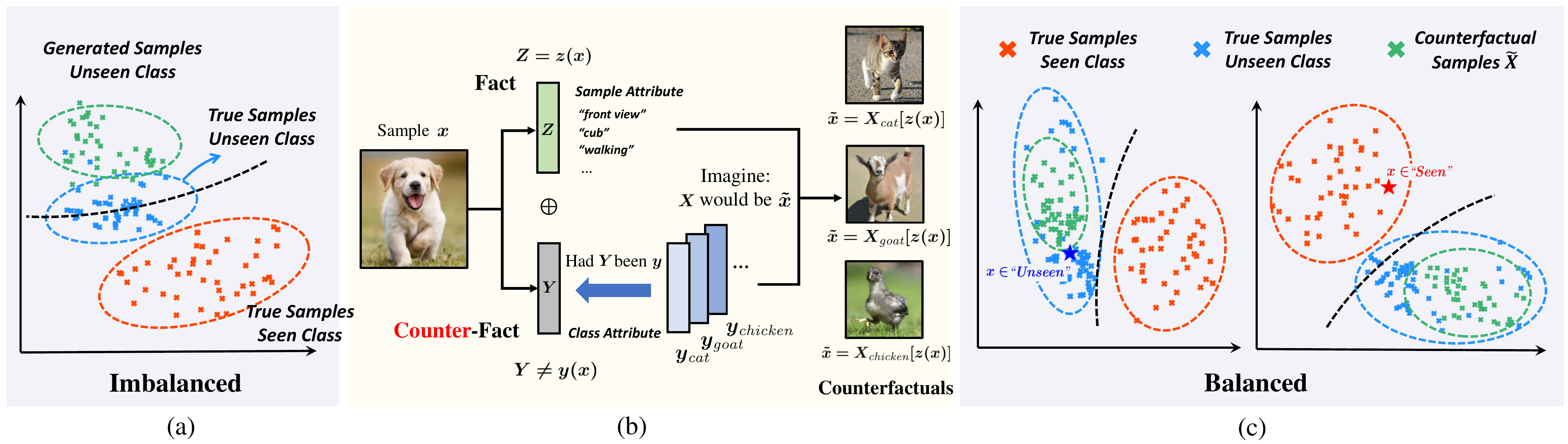}
         \phantomcaption
         \label{fig:2a}
    \end{subfigure}
    \begin{subfigure}[t]{0\textwidth} 
         \includegraphics[width=\textwidth]{example-image-b}
         \phantomcaption
         \label{fig:2b}   
    \end{subfigure}
    \begin{subfigure}[t]{0\textwidth} 
         \includegraphics[width=\textwidth]{example-image-b}
         \phantomcaption
         \label{fig:2c}   
    \end{subfigure}
    \vspace*{-9mm}
    \caption{(a) t-SNE~\cite{maaten2008visualizing} plot of the CUB~\cite{WelinderEtal2010} samples in ZSL using a conventional unseen generation method~\cite{narayan2020latent}, where a single pair of seen- and unseen-class is shown to avoid clutter. Due to the out-of-distribution generation (green), the decision boundary (black dashed lines) is imbalanced between the true seen (red) and unseen (blue) samples. (b) Illustration of our counterfactual generation. (c) t-SNE plot of the CUB using our counterfactual generation. The decision boundary is balanced. See Figure~\ref{fig:4} for more examples of OSR.}
    \vspace{-5mm}
    \label{fig:2}
\end{figure*}

Not surprisingly, the above assumption is hardly valid in practice. As the model only sees the seen-classes in training, it will inevitably cater to the seen idiosyncrasies, and thus result in an unrealistic imagination of the unseen world. 
%
Figure~\ref{fig:2a} illustrates that the samples, generated from the class attribute of an unseen-class, do not lie in the sample domain between the ground-truth seen and unseen, \ie, they resemble neither the seen nor the unseen. As a result, the seen/unseen boundary learned from the generated unseen and the true seen samples is imbalanced. Such imbalanced classification increases the recall of the seen-class by sacrificing that of the unseen. Interestingly, we find that all the existing ZSL methods suffer from this imbalance: the seen accuracy is much higher than the unseen (\textit{cf}. Table~\ref{tab:3}). 

Astute readers may intuitively realize that it is all about \emph{disentanglement}: if every attribute is disentangled, any unseen combination will be sensible. However, it is well-known that learning disentangled features is difficult, or even impossible without proper supervision~\cite{locatello2019challenging}. In this paper, we propose another way around: \textbf{Counterfactual Inference}~\cite{pearl2016causal}, which does not require the disentanglement for the class attributes. As illustrated in Figure~\ref{fig:2b}, denote $X$ as the sample variable, which can be encoded into the \emph{sample} attributes $Z = z(X=x)$ (\eg, ``front view'') and \emph{class} attributes $Y=y(X=x)$. The counterfactual~\cite{pearl2009causality}: $\tilde{x}={X_y}[z(x)]$ of sample $x$, is read as:
\begin{center}
\framebox[1.5\width]{\Longstack[c]{$X$ would be $\tilde{x}$, had $Y$ been $y$,\\
given the fact that $Z = z(X = x)$.}} \par
\end{center}
Note that the given ``fact'' is $Z = z(x)$ and the ``counter-fact'' (what if) is  $Y\neq y(x)$, indicating that the encoded $y(x)$ --- also an observed fact --- clashes with the counter-fact $y$. The key difference between the conventional unseen generation and the counterfactual is that: the former is purely based on the \textbf{sample-agnostic} class attributes $y$ (or together with Guassian prior $z$~\cite{xian2018feature}), while the latter is also grounded by the \textbf{sample-specific} attributes $z(x)$.

So, why sample-specific? 
The imagination of an unseen-class is indeed not necessary to start from scratch, \ie, purely from the class definition $Y$, where the effect of some attributes may be lost due to entanglement~\cite{chen2018zero}; instead, it should also start from the observed fact $Z = z(x)$, which can make up those entangled attributes. In fact, such grounded generation simulates how we humans imagine an unseen sample~\cite{pearl2018book}: we imagine a ``dinosaur'' (class attributes) based on its fossil (sample attributes). More formally, by disentangling the two groups: class attribute $Y$ and sample $Z$, we can use the theorem of \textbf{Counterfactual Faithfulness} (Section~\ref{sec:3.3} \&~\ref{sec:3.4}) to guarantee that the counterfactual distribution is coherent with the ground-truth seen/unseen distribution. As shown in Figure~\ref{fig:2c} (left), the unseen-class generation grounded by a sample is in the true unseen domain, leading to a more balanced decision boundary. It is worth noting that the group disentanglement between $Z$ and $Y$ is much more relaxed and thus more approachable than the full disentanglement~\cite{besserve2020counterfactual,besserve2018group}, \ie, the attributes within each group are not required to be disentangled. In Section~\ref{sec:3.4}, we design a training procedure to achieve this.

We propose a counterfactual framework for ZSL and OSR, because they are both underpinned by the generalization to unseen-classes. As summarized in Figure~\ref{fig:1}, the counterfactual is powered by a Generative Causal Model (Section~\ref{sec:3.2}). The counterfactual faithfulness allows us to use the \textbf{Consistency Rule}: if the counter-fact $y$ is indeed the underlying ground-truth, the counterfactual $\tilde{x}$ equals to the factual $x$. Therefore, we can use the rule as a seen/unseen binary classifier by varying $y$ across the seen/unseen class attributes (Section~\ref{sec:3.3}): If a sample is seen, we can apply the conventional supervised learning classifier; otherwise, for ZSL, we apply the conventional ZSL classifier, and for OSR, we reject it.


To the best of our knowledge, the proposed counterfactual framework is the first to provide a theoretical ground for balancing and improving the seen/unseen classification. In particular, we show that the quality of disentangling $Z$ and $Y$ is the key bottleneck, so it is a potential future direction for ZSL/OSR~\cite{suter2019robustly,higgins2018towards,parascandolo2018learning,besserve2020theory}.
Our method can serve as an unseen/seen binary classifier, which can plug-and-play and boost existing ZSL/OSR methods to achieve new state-of-the-arts (Section~\ref{sec:4}).
%
\section{Related Work}
\label{sec:2}
\textbf{ZSL} is usually provided with an auxiliary set of class attributes to describe each seen- and unseen-class~\cite{lampert2009learning,xian2018zero}. Therefore, ZSL can be approached by either inferring a sample's attribute and finding the closest match in the attribute space~\cite{frome2013devise,akata2015label,chen2018zero, jiang2019transferable}, or generating features using the attributes and matching in the feature space~\cite{xian2018feature,li2019leveraging,narayan2020latent, pambala2020generative}. 
\textbf{OSR} is a more challenging open environment setting with no information on the unseen-classes~\cite{scheirer2013toward,scheirer2014probability}, and the goal is to build a classifier for seen-classes that additionally rejects unseen-class samples as outliers. The inference methods in OSR calibrate a discriminative model by adjusting the classification logits~\cite{bendale2016towards,zongyuan2017generative}, and the generation methods estimate the seen-class density and reject test samples in the low-density area as outliers~\cite{oza2019c2ae,sun2020conditional}. \textbf{Out-Of-Distribution (OOD)} detection~\cite{grathwohl2019your,arjovsky2019invariant,liu2020energy} also focuses on unseen detection, where seen and unseen samples are usually from different domains~\cite{bulusu2020anomalous}. However, in OOD, the unseen-class information is available. In particular, some works employ OOD techniques in ZSL to distinguish seen- and unseen-classes~\cite{mandal2019out,chen2020boundary}. Our work is based on \textbf{Causal Inference}~\cite{pearl2009causality}, which has shown promising results in various computer vision tasks~\cite{zhang2020causal, wang2020visual, yang2020deconfounded} including few-shot classification~\cite{yue2020interventional}, long-tailed classification~\cite{kaihua2020long} and incremental learning~\cite{hu2021distilling}. In particular, a recent paper~\cite{atzmon2020causal} uses causal intervention on ZSL. However, it builds on a restrictive setting where class attributes are fully disentangled (\eg, shape and color). Our work uses \emph{counterfactual inference} and relies on the much-relaxed group disentanglement~\cite{besserve2018group}, allowing us to outperform on complex benchmark datasets.

\section{Approach}

\label{sec:3}

\subsection{Problem Definitions}

\label{sec:3.1}
We train a model using a labelled dataset $\mathcal{D}=\{\mathbf{x}_i,l_i\}_{i=1}^N$ on seen-classes $\mathcal{S}$, to recognize samples from both $\mathcal{S}$ and unseen-classes $\mathcal{U}$, where $\mathbf{x}_i\in\mathcal{X}\subset \mathbb{R}^d$ is the $d$-dimensional feature vector of the $i$-th sample and $l_i \in \mathcal{S}$ is its corresponding label. We assume that the samples of $\mathcal{S}$ and $\mathcal{U}$ are embedded in the same feature space $\mathcal{X}$, whose ambient space is the RGB color space or the network feature space provided by the dataset curator~\cite{xian2018zero}.

\noindent\textbf{Zero-Shot Learning (ZSL)}. It includes two settings: 1) \textit{Conventional ZSL}, where the model is only evaluated on $\mathcal{U}$, and 2) \textit{Generalized ZSL}, where the model is evaluated on $\mathcal{S} \cup \mathcal{U}$. A common practice~\cite{chao2016empirical} is to use an additional set of class attribute vectors $\mathcal{Y}_S$ and $\mathcal{Y}_U$ to describe the seen- and unseen-classes, respectively. Compared to the one-hot label embeddings, these attributes can be considered as dense label embeddings~\cite{schwartz2019baby,munro2020multi}. When the context is clear, we refer to ZSL as Generalized ZSL.

\noindent\textbf{Open-Set Recognition (OSR)}. It is evaluated on both $\mathcal{S}$ and $\mathcal{U}$. Compared to ZSL, $\mathcal{U}$ in OSR is marked as ``unknown''. Instead of using dense labels, each seen-class is described by the one-hot embedding of $K$ dimensions, where $K$ is the number of seen-classes. The one-hot embeddings are considered as the seen-class attributes set $\mathcal{Y}_S$. Since OSR is evaluated in an open environment, there is no unseen-class attributes set $\mathcal{Y}_U$.

\subsection{Generative Causal Model}
\label{sec:3.2}

\begin{wrapfigure}{r}{0.10\textwidth}
    \centering
    \includegraphics[width=.10\textwidth]{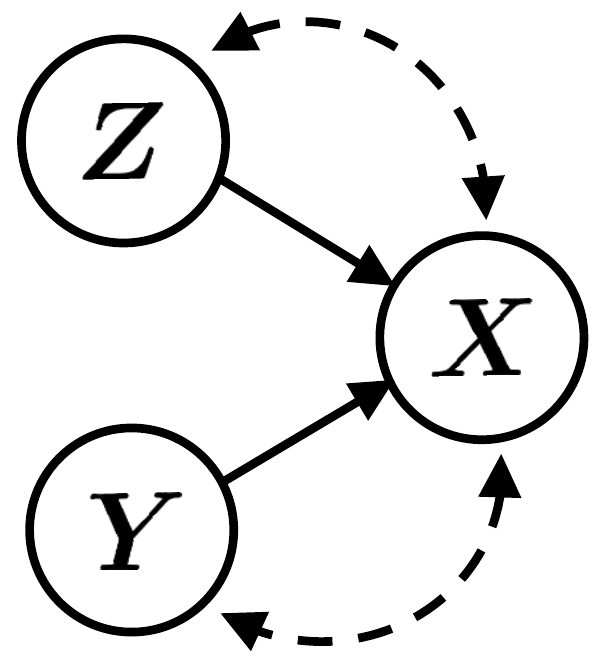}
    \caption{Our GCM for ZSL and OSR.}
    \label{fig:3}
    \vspace{-3mm}
\end{wrapfigure}

Our assumption is that both ZSL and OSR follow a Generative Causal Model (GCM)~\cite{pearl2009causality} shown in Figure~\ref{fig:3}, where the \emph{class attribute} $Y$ and the \emph{sample attribute} $Z$  jointly determine the observed image feature $X$. In general, the generative causal process $Z\rightarrow X, Y\rightarrow X$ can be confounded, represented as the dashed links in Figure~\ref{fig:3}. 
One can remove the confounders through more elaborate data collection~\cite{xian2018zero} and experimental design~\cite{imai2013experimental}. In this paper, we follow the conventions in ZSL and OSR to ignore the confounders~\cite{li2019leveraging,oza2019c2ae}. Our GCM entails a generation and inference process. Specifically, given $Z$ and $Y$, we can generate $X$ by sampling from the conditional distribution $P_\theta(X|Z,Y)$. While given $X$, we can infer $Z$ and $Y$ through the posterior $Q_\phi(Z|X)$ and $Q_\psi(Y|X)$. Therefore, GCM can support both generative and inference-based methods. As the former prevails over the latter in literature, we focus on it in this paper.

In ZSL, $Y$ takes values fom the dense labels set $\mathcal{Y}_S\cup\mathcal{Y}_U$. Generative ZSL methods~\cite{xian2018feature,li2019leveraging} aim to learn $P_\theta(X|Z,Y)$. In testing, they generate unseen-class samples $X$ by using a Gaussian prior as $Z$ and the attributes $Y$ in $\mathcal{Y}_U$. Then, the generated unseen samples and the seen $\mathcal{D}$ are used to train a classifier to recognize both $\mathcal{S}$ and $\mathcal{U}$. In OSR, the value of $Y$ is from the one-hot embedding set $\mathcal{Y}_S$. Generative OSR methods~\cite{oza2019c2ae,sun2020conditional} first infer $Y$ from a test sample $\mathbf{x}$ by sampling $\mathbf{y} \sim Q_\psi(Y|X=\mathbf{x})$. Then, the inferred $\mathbf{y}$ and Gaussian noise $\mathbf{z}$ are used to generate $\mathbf{x}'$ from $P_\theta(X|Z=\mathbf{z},Y=\mathbf{y})$. Finally, the sample is marked as ``unknown'' if $\mathbf{x}$ is dissimilar to $\mathbf{x}'$ subject to a specified threshold.

Note that all the above generation methods adopt a prior Gaussian noise for $Z$, which is not sample-specific. As $Y$ is inevitably entangled (examples in Appendix), there is no mechanism to make up for the missing sample attribute effect during generation, so, the generated unseen-class samples will be unrealistic and lie outside $\mathcal{X}$, rendering the decision rule trained on the seen samples inapplicable.

\subsection{Counterfactual Generation and Inference}
\label{sec:3.3}

\begin{figure}
    \centering
    \footnotesize
    
    \begin{subfigure}[t]{1.0\linewidth}
        \centering
        \includegraphics[width=0.9\textwidth]{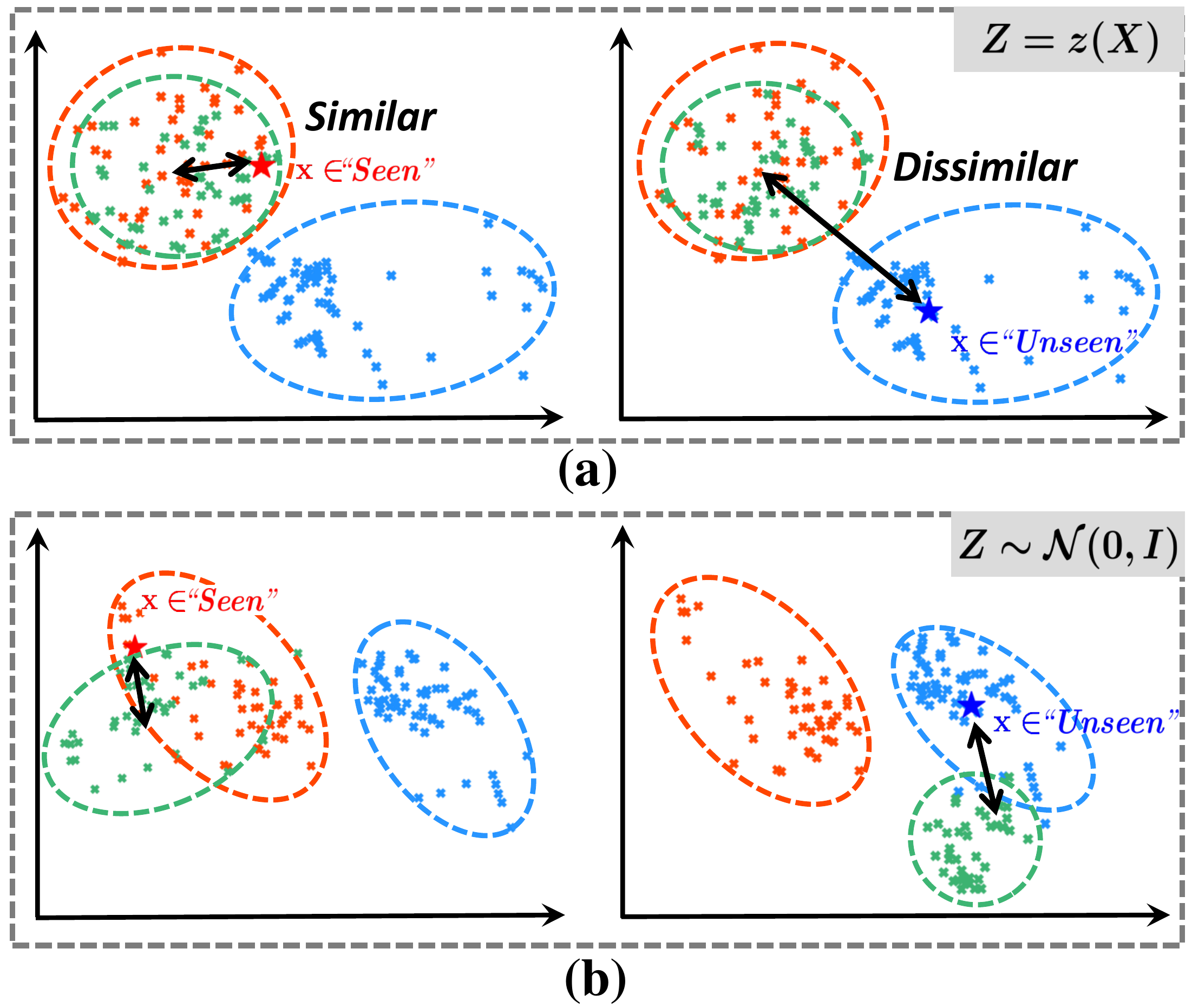}
        \phantomcaption
        \label{fig:4a}   
    \end{subfigure}
    \begin{subfigure}[t]{0\textwidth} 
         \includegraphics[width=\textwidth]{example-image-b}
         \phantomcaption
         \label{fig:4b}   
    \end{subfigure}
    \vspace*{-8mm}
    \caption{OSR using (a) our counterfactual framework ($Z$=$z(X)$); (b) existing generative approach ($Z \sim \mathcal{N}(0,I)$) when the test sample is from seen-classes (left) and unseen-classes (right). \textcolor{red}{Red dashed line} denotes the true samples from the seen-class and \textcolor{blue}{blue dash line} as those from the unseen-class. \textcolor{green}{Green dash line} denotes generated samples using the seen-class attribute. \textcolor{red}{$\star$} and \textcolor{blue}{$\star$} 
   denote the samples from seen and unseen class, respectively.}
    \label{fig:4}
    \vspace*{-5mm}
\end{figure}

Given a sample $\mathbf{x}$, we use the GCM to generate counterfactual samples $\tilde{\mathbf{x}}=X_\mathbf{y}[z(\mathbf{x})]$ following the three steps of computing counterfactual~\cite{pearl2016causal}:

\noindent\textbf{Abduction}-\textit{``given the fact that $Z=z(\mathbf{x})$''}. We solve for the endogenous sample attribute $z(\mathbf{x})$ given the evidence $X=\mathbf{x}$. In our GCM, we can sample from the posterior $z(\mathbf{x})\sim Q_{\phi}(Z|X=\mathbf{x})$.

\noindent\textbf{Action}-\textit{``had Y been y''}. Here $\mathbf{y}\in \mathcal{Y}_S \cup \mathcal{Y}_U$ is the intervention target of $Y$. Note that the class attribute $Y$ of this sample can be inferred from $y(\mathbf{x}) \sim Q_\psi(Y|X=\mathbf{x})$, but in this step, we intervene on $Y$ by discarding the inferred value and setting $Y$ as $\mathbf{y}$, which may be different from $y(\mathbf{x})$.
 
\noindent\textbf{Prediction}-\textit{``X would be $\tilde{\mathbf{x}}$''}. Conditioning on the inferred $Z=z(\mathbf{x})$ (fact) and the intervention target $Y=\mathbf{y}$ (counter-fact), we can generate the counterfactual sample $\tilde{\mathbf{x}}$ from $P_{\theta}(X|Z=z(\mathbf{x}), Y=\mathbf{y})$. Notice that the feature imagination from class attribute $Y=\mathbf{y}$ is now grounded by the sample attribute $Z=z(\mathbf{x})$.

We want that the above counterfactual sample lies in the true distribution of the seen or unseen samples. Formally, such property is defined as below:

\noindent\textbf{Definition (Counterfactual Faithfulness)}. \textit{Given $\mathbf{x} \in \mathcal{X}$, the counterfactual generation $\tilde{\mathbf{x}}$ using a GCM is faithful whenever $\tilde{\mathbf{x}} \in \mathcal{X}$.}

In fact, the failure of existing methods, as mentioned in Section~\ref{sec:3.2}, is because the counterfactual faithfulness does not hold; however, it holds for our generation. We delay the justification to  Section~\ref{sec:3.4}. It assures that any distance metric in $\mathcal{X}$ is applicable for both $\mathbf{x}$ and its counterfactual generation $\tilde{\mathbf{x}}$. This allows us to build a binary classifier on seen/unseen by applying the \textbf{Consistency Rule}:
\begin{equation}
    y^*(\mathbf{x})=\mathbf{y} \Longrightarrow X_\mathbf{y}[z(\mathbf{x})]=\mathbf{x},
\end{equation}
where $y^*(\mathbf{x})$ is the (unobserved) ground-truth class attribute of $\mathbf{x}$ and $X_\mathbf{y}[z(\mathbf{x})]$ is the generated counterfactual with this ground-truth. Our binary classification strategy is based on the equivalent \textit{contraposition} of the rule, which states that \textit{if $\mathbf{x}$ is dissimilar to $X_\mathbf{y}[z(\mathbf{x})]$, the ground-truth attribute of $\mathbf{x}$ cannot be $\mathbf{y}$}:
\begin{equation}
    X_\mathbf{y}[z(\mathbf{x})] \neq \mathbf{x} \Longrightarrow y^*(\mathbf{x}) \neq \mathbf{y}.
\end{equation}
Thanks to the counterfactual faithfulness, the dissimilarity can be measured by any distance metric defined in $\mathcal{X}$ (\eg, Euclidean distance). Figure~\ref{fig:4} uses the OSR task to show the benefit of our counterfactual approach grounded by $Z=z(X)$. The class-agnostic $Z$ in the existing methods cannot make up for the attributes that entangled in $Y$. This leads to non-faithful generation on unseen-class samples, and the distance can hardly tell apart the seen and unseen-class samples (Figure~\ref{fig:4b}). In contrast, our approach achieves counterfactual faithfulness and thus the distance remains discriminative (Figure~\ref{fig:4a}). Next, we detail the binary inference rule for both ZSL and OSR.

\noindent\textbf{Inference in ZSL}. Since ZSL is evaluated in a closed environment, \ie, the set of unseen-classe attributes $\mathcal{Y}_U$ is known in testing, we use the contraposition---if feature $\mathbf{x}$ is \emph{dissimilar} to counterfactual generations from the \textit{unseen}-classes, $\mathbf{x}$ belongs to \textit{seen}. For example in Figure~\ref{fig:2c}, conditioning on a seen-class sample $\mathbf{x}$, the counterfactual generations using the unseen-class attribute are indeed dissimilar to $\mathbf{x}$ as they lie in the opposite side of the classifier decision boundary. Specifically, we generate a set of counterfactual features $\tilde{X}$ of the \emph{unseen}-classes by using the inferred $Z$ from $Q_\phi(Z|X=\mathbf{x})$ and intervening $Y$ with the ground-truth attributes from $\mathcal{Y}^U$. Using the counterfactual set $\tilde{X}$ of the unseen-classes and $\mathcal{D}$ of the seen-classes, we train a multi-label classifier whose vocabulary is $\mathcal{S}\cup\mathcal{U}$. Denote the mean-pooling of the top-$K$ classifier probabilities among seen- and unseen-classes as $S^K$ and $U^K$, respectively. The binary seen/unseen label $b(\mathbf{x})$ is given by:
\begin{equation}
    b(\mathbf{x})=
    \begin{cases}
    \text{seen},& \text{if } U^K < S^K\\
    \text{unseen},& \text{otherwise}
    \end{cases}
\end{equation}

\noindent\textbf{Inference in OSR}. Since OSR is in an open environment, \ie, there could be infinite number of unseen-classes, it is impossible to generate unseen-class counterfactuals. Therefore, we use the contraposition the other way round---if $\mathbf{x}$ is \emph{dissimilar} to the counterfactual generations of the \emph{seen}-classes, $\mathbf{x}$ belongs to the \emph{unseen}. This is shown in Figure~\ref{fig:4a}, where the unseen-class sample (right), compared to the seen-class sample (left), is much more dissimilar to the seen-class counterfactuals. Hence by thresholding the dissimilarity, we can classify both samples correctly. Specifically, we generate a set of counterfactual features $\tilde{X}$ on \emph{seen}-classes by setting $Y$ as the one-hot embeddings in $\mathcal{Y}_S$, and we calculate the minimum Euclidean distance between $\mathbf{x}$ and each $\tilde{\mathbf{x}} \in \tilde{X}$, denoted as $d_{min}$. If $d_{min}$ is larger than a threshold $\tau$, the sample is dissimilar to the seen-classes counterfactuals. The binary label $b(\mathbf{x})$ is given by:
\begin{equation}
    b(\mathbf{x})=
    \begin{cases}
    \text{unseen},& \text{if } d_{min} > \tau\\
    \text{seen},& \text{otherwise}
    \end{cases}
\end{equation}

\noindent\textbf{Two-Stage Inference}. As shown in Figure~\ref{fig:1}, after the stage-one binary classification, at the second stage, the predicted seen-classes samples can be classified by using the standard supervised classifier, and the predicted unseen-classes samples can be fed into any Conventional ZSL algorithms in ZSL, or rejected as outliers in OSR. The detailed implementation is in Section~\ref{sec:4.2}.

\begin{figure}
    \centering
    \footnotesize
    
    \includegraphics[width=.9\linewidth]{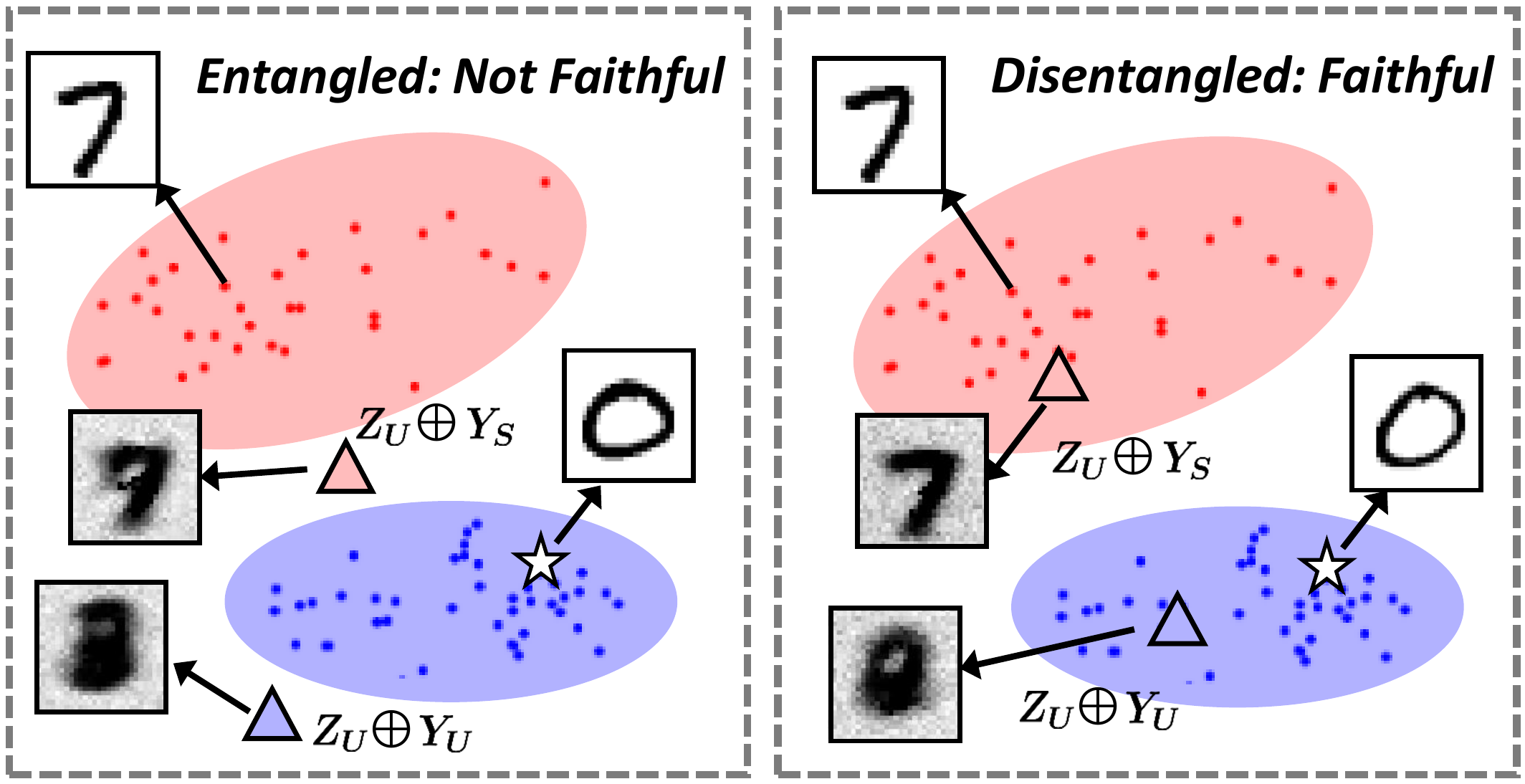}
    \vspace*{-2mm}
    
    \caption{Counterfactual generation conditioned on an unseen sample (denoted as the star). $Z_U \oplus Y_S$ and $Z_U \oplus Y_U$ represent counterfactual images conditioned on the unseen sample, using seen and unseen attribute respectively.}
    \label{fig:5}
    \vspace*{-4mm}
\end{figure}

\subsection{Counterfactual-Faithful Training}
\label{sec:3.4}

Our framework centers on the counterfactual-faithful generation, which can be guaranteed by the following theorem (proved as a corollary of~\cite{besserve2020counterfactual}, Appendix):

\noindent\textbf{Theorem}. \textit{The counterfactual generation $X_\mathbf{y}[z(\mathbf{x})]$ is faithful if and only if the sample attribute $Z$ and class attribute $Y$ are group disentangled.}

Figure~\ref{fig:5} shows the difference between the generations with and without the group disentanglement. Therefore, to achieve counterfactual faithful generation, the full disentanglement among each dimension of $Y$ and $Z$, which may be impossible in general~\cite{locatello2019challenging}, is relaxed to group disentanglement between $Y$ and $Z$, which can be more easily addressed by the following specially designed training objective:
\begin{equation}
    \min_{\theta,\phi} \mathcal{L}_Z + \nu \mathcal{L}_Y + \max_{\omega} \rho \mathcal{L}_F,
    \vspace{-3mm}
\end{equation}
where $\nu,\rho$ are a trade-off parameters. All three losses are designed to achieve the counterfactual faithfulness, which assures the distance metric of the consistency rule. Next, we detail each of them.

\noindent\textbf{Disentangling $Z$ from $Y$}. We minimize the $\beta$-VAE~\cite{Higgins2017betaVAELB} loss $\mathcal{L}_Z$ given by:
\begin{equation}
\begin{split}
    \mathcal{L}_{Z} = &-\mathbb{E}_{Q_{\phi}(Z|X)}\left[ P_{\theta}(X\mid Z,Y) \right] \\
    &+ \beta D_{KL}\left(Q_{\phi}(Z\mid X) \parallel P(Z)\right),
\end{split}
\end{equation}
where $D_{KL}$ denotes KL-divergence, $P_\theta(X|Z,Y)$ and $Q_\phi(Z|X)$ are implemented using the Deep Gaussian family~\cite{kingma2014ICLR}, and the prior $P(Z)$ is set to the isotropic Gaussian distribution. Compared to the standard VAE objective, $\beta$-VAE re-weighs the KL divergence term by a factor of $\beta$ ($\beta > 1$). This is shown to be highly effective in learning a disentangled sample attribute $Z$~\cite{Higgins2017betaVAELB,suter2019robustly}. Intuitively, by placing a strict constraint for $Z$ as a whole to follow the endogenous prior $P(Z)$, the value of $Z$ becomes unaffected by the distribution of $Y$, \ie, $Z$ is disentangled from $Y$.

\noindent\textbf{Disentangling $Y$ from $Z$}. However, the above regularization cannot guarantee that the GCM correctly uses the class attribute $Y$ during generation. For example, recent GAN models~\cite{zhu2017unpaired,karras2019style} can generate a large variety of photo-realistic images by only using $Z$. In fact, as shown in recent literature~\cite{besserve2020theory}, it is possible for an over-parameterized model $P_\theta(X|Z,Y)$ to ignore $Y$ and use purely $Z$ to generate $X$, leading to non-faithful generations. This is because the information in $Y$ might be \emph{fully contained} in $Z$. Therefore, it is necessary to additionally disentangle $Y$ from $Z$. Specifically, given a training sample $\mathbf{x}$, its ground-truth attribute $\mathbf{y}$ and its sample attribute $z(\mathbf{x})$ sampled from $Q_\phi(Z|X=\mathbf{x})$, we require $\mathbf{x}$ to be close to $\mathbf{x}_{\mathbf{y}} = X_{\mathbf{y}}[z(\mathbf{x})]$, but far away from the counterfactual generations in the set $\tilde{X} = \{ X_{\mathbf{y}'}[z(\mathbf{x})] \mid \mathbf{y}' \in \mathcal{Y}^S \wedge \mathbf{y}' \neq \mathbf{y}\}$. We use a contrastive loss as:
\begin{equation}
    \mathcal{L}_Y = - log \frac{exp \left(-dist(\mathbf{x}, \mathbf{x}_{\mathbf{y}} ) \right)} {\sum_{\mathbf{x}'\in \tilde{X} \cup \{\mathbf{x}_{\mathbf{y}}\}} exp\left( -dist(\mathbf{x},\mathbf{x}')\right)},
\end{equation}
where $dist$ denotes Euclidean distance. Intuitively, this loss actively intervenes the class attribute $Y$, given fixed sample attributes $Z$, and therefore maximizes the sample difference (\eg, in terms of constrastiveness) before and after the intervention, which complies with the recent definition of causal disentanglement~\cite{higgins2018towards}. Therefore, $\mathcal{L}_Y$ can disentangles $Y$ from $Z$.

\noindent\textbf{Further Disentangling by Faithfulness}. Note that the faithfulness $\tilde{\mathbf{x}}\in\mathcal{X}$ is a necessary condition for disentanglement, so we can also use it as an objective. Note that the VAE objective optimizes a lower bound of the likelihood $P(X)$, where the bound looseness undermines the faithfulness. To address this problem, we additionally use the Wasserstein GAN (WGAN)~\cite{wasserstein2017martin} loss. Specifically, we train a discriminator $D(X,Y)$ parameterized by $\omega$ that outputs a real value, indicating if feature $X$ is realistic conditioned on $Y$ (larger is more realistic). Given a feature $\mathbf{x}$ with attribute $\mathbf{y}$, we can generate the counterfactual $\mathbf{x}'$ using $z(\mathbf{x})$ and $\mathbf{y}$. The discriminator is trained to mark $\mathbf{x}$ as real (large $D(\mathbf{x},\mathbf{y})$) and $\mathbf{x}'$ as unreal (small $D(\mathbf{x}',\mathbf{y})$). Specifically, the WGAN loss $\mathcal{L}_F$ is given by:
\begin{equation}
\begin{split}
    \mathcal{L}_F = &\mathbb{E}[D(\mathbf{x},\mathbf{y})] - \mathbb{E}[D(\mathbf{x}',\mathbf{y})] \\
    &- \lambda \mathbb{E}\left[(\norm{\nabla_{\hat{\mathbf{x}}} D(\hat{\mathbf{x}}, \mathbf{y})}_2 - 1 )^2 \right],
\end{split}
\end{equation}
where $\lambda$ is a penalty term, $\hat{\mathbf{x}} = \alpha \mathbf{x} + (1-\alpha) \mathbf{x'}$ with $\alpha$ sampled from the uniform distribution defined on $[0,1]$. By training the generation process $P_\theta(X|Z,Y)$ and the discriminator in an adversarial fashion, the GCM is regularized to generate $\mathbf{x}'$ that is similar to $\mathbf{x}$ in the same distribution, \ie, counterfactual-faithful for seen-class samples.


\section{Experiments}
\label{sec:4}

\subsection{Datasets}
\label{sec:4.1}
\noindent\textbf{ZSL}. 
We evaluated our method on standard benchmark datasets: Caltech-UCSD-Birds 200-2011 (\textit{CUB})~\cite{WelinderEtal2010}, \textit{SUN}~\cite{xiao2010sun}, Animals with Attributes 2 (\textit{AWA2})~\cite{xian2018zero} and attribute Pascal and Yahoo (\textit{aPY})~\cite{farhadi2009describing}. In particular, we followed the unseen/seen split in the \emph{Proposed Split (PS) V2.0}~\cite{xian2018zero}, recently released to fix a test-data-leaking bug in the original PS. The granularity, total \#images, \#seen-classes ($|\mathcal{S}|$) and \#unseen-classes ($|\mathcal{U}|$) are given in Table~\ref{tab:1}.

\noindent\textbf{OSR}.
We used the standard evaluation datasets: \textit{MNIST}~\cite{lecun2010mnist}, \textit{SVHN}~\cite{netzer2011reading}, \textit{CIFAR10}~\cite{krizhevsky2009learning} and \textit{CIFAR100}~\cite{krizhevsky2009learning}. The image size, \#classes and \# images in train/test split of the datasets are given in Table~\ref{tab:2}. Following the standard benchmark~\cite{neal2018open,yoshihashi2019classification} in OSR, we split MNIST, SVHN and CIFAR10 into 6 seen-classes and 4 unseen-classes, and construct two additional datasets CIFAR+10 (\textit{C+10}) and CIFAR+50 (\textit{C+50}), where 4 non-animal classes in CIFAR10 are used as seen-classes, while additional 10 and 50 animal classes from CIFAR100 are used as unseen-classes.

\begin{table}[h!]
\centering
\captionsetup{font=footnotesize,labelfont=footnotesize,skip=5pt}
\scalebox{0.8}{
\begin{tabular}{p{2cm} p{2cm}<{\centering}p{1.2cm}<{\centering}p{1.2cm}<{\centering}p{1.2cm}<{\centering}}
\hline\hline
\multicolumn{1}{c}{\large{Dataset}} & Granularity & Total & $|\mathcal{S}|$ & $|\mathcal{U}|$\\
\hline
CUB~\cite{WelinderEtal2010} & Fine & 11,788 & 150 & 50 \\
SUN~\cite{xiao2010sun} & Fine & 14,340 & 645 & 72 \\
AWA2~\cite{xian2018zero} & Coarse & 37,322 & 40 & 10 \\
aPY~\cite{farhadi2009describing} & Coarse & 12,051 & 20 & 12\\
\hline \hline
\end{tabular}}
\caption{Information on ZSL datasets.}
\label{tab:1}
\vspace{-0.3cm}
\end{table}

\begin{table}[h!]
\centering
\captionsetup{font=footnotesize,labelfont=footnotesize,skip=5pt}
\scalebox{0.8}{
\begin{tabular}{p{2cm} p{2cm}<{\centering}p{1.2cm}<{\centering}p{1.2cm}<{\centering}p{1.2cm}<{\centering}}
\hline\hline
\multicolumn{1}{c}{\large{Dataset}} & Image Size & Classes & Train & Test\\
\hline
MNIST~\cite{lecun2010mnist} & 28$\times$28 & 10 & 60,000 & 10,000 \\
SVHN~\cite{netzer2011reading} & 32$\times$32 & 10 & 73,257 & 26,032 \\
CIFAR10~\cite{krizhevsky2009learning} & 32$\times$32 & 10 & 50,000 & 10,000 \\
CIFAR100~\cite{krizhevsky2009learning} & 32$\times$32 & 100 & 50,000 & 10,000\\
\hline \hline
\end{tabular}}
\caption{Information on OSR datasets.}
\label{tab:2}
\vspace{-0.3cm}
\end{table}

\begin{table*}[t!]
\centering
\captionsetup{font=footnotesize,labelfont=footnotesize,skip=5pt}
\scalebox{0.8}{
\begin{tabular}{p{0.8cm}p{2.7cm} p{0.2cm}<{\centering}p{0.6cm}<{\centering}p{0.6cm}<{\centering}p{0.6cm}<{\centering}p{0.2cm}<{\centering}p{0.6cm}<{\centering}p{0.6cm}<{\centering}p{0.6cm}<{\centering}p{0.2cm}<{\centering} p{0.6cm}<{\centering}p{0.6cm}<{\centering}p{0.6cm}<{\centering}p{0.2cm}<{\centering} p{0.6cm}<{\centering}p{0.6cm}<{\centering}p{0.6cm}<{\centering}}
\hline\hline
\multirow{2}{*}{}  & \multicolumn{1}{c}{\multirow{2}{*}{\large{Method}}} &     & \multicolumn{3}{c}{\textbf{CUB}} & &              \multicolumn{3}{c}{\textbf{AWA2}} & & \multicolumn{3}{c}{\textbf{SUN}} & & \multicolumn{3}{c}{\textbf{aPY}} \\ \cmidrule(lr){4-6}\cmidrule(lr){8-10}\cmidrule(lr){12-14}\cmidrule(lr){16-18}
& \multicolumn{1}{c}{}    &   & $U$   & $S$    & $H$      &    & $U$ & $S$   & $H$  &   & $U$   & $S$    & $H$ & & $U$   & $S$    & $H$  \\ \hline
\multicolumn{1}{c}{\multirow{4}{*}{\rotatebox{90}{\large{Inf.}}}} &                 ALE$^\dagger$~\cite{akata2015label} & & 23.7 & 62.8 & 34.4  &  & 14.0 & 81.8 & 23.9  & & 21.8         & 33.1 & 26.3 & & 4.6 & 73.7 & 8.7 \\
    & DEVISE$^\dagger$~\cite{frome2013devise}  &  & 23.8 & 53.0 & 32.8  &  & 17.1 & 74.7 & 27.8      & & 16.9 & 27.4 & 20.9 & & 3.5 & 78.4 & 6.7 \\
    & LATEM$^\dagger$~\cite{xian2016latent} & & 15.2 & 57.3 & 24.0 & & 11.5  & 77.3 & 20.0 &     & 14.7 & 28.8 & 19.5 & & 1.3 & 71.4 & 2.6   \\
    & RelationNet~\cite{sung2018learning} & & 36.3 & \textbf{63.8} & 46.3 &  & 22.1 & \textbf{91.4} & 35.5 & & 15.8  & 25.5 & 19.6 & & 11.5 & \textbf{80.7} & 20.2 \\
\hline
\multicolumn{1}{c}{\multirow{5}{*}{\rotatebox{90}{\large{Gen.}}}}  &               GDAN~\cite{huang2019generative} & & 35.0 & 28.7 & 31.6 & & 26.0 & 78.5 & 39.1 & & 38.2 & 19.8 & 26.1 & & 29.0 & 63.7 & 39.9 \\
     & CADA-VAE~\cite{schonfeld2019generalized} & & 50.3 & 56.1 & 53.0 & & 55.4 & 76.1 & 64.0 & & 43.6 & 36.4 & 39.7 & & 34.0 & 54.2 & 41.7 \\
     & LisGAN~\cite{li2019leveraging} & & 44.9 & 59.3 & 51.1 & & 53.1 & 68.8 & 60.0 & & 41.9 & 37.8 & 39.8 & & 33.2 & 56.9 & 41.9 \\
     & TF-VAEGAN~\cite{narayan2020latent} & & 50.7 & 62.5 & 56.0 & & 52.5 & 82.4 & 64.1  & & 41.0 & \textbf{39.1} & 40.0 & & 31.7 & 61.5 & 41.8 \\
     & \textbf{GCM-CF (Ours)} & & \cellcolor{mygray}\textbf{61.0} & \cellcolor{mygray}59.7 & \cellcolor{mygray}\textbf{60.3} & \cellcolor{mygray} & \cellcolor{mygray}\textbf{60.4} &  \cellcolor{mygray}75.1 & \cellcolor{mygray}\textbf{67.0} & \cellcolor{mygray}  & \cellcolor{mygray}\textbf{47.9} & \cellcolor{mygray}37.8 & \cellcolor{mygray}\textbf{42.2} & \cellcolor{mygray} & \cellcolor{mygray}\textbf{37.1} & \cellcolor{mygray}56.8 & \cellcolor{mygray}\textbf{44.9} \\ \hline \hline
\end{tabular}}
\caption{ZSL Accuracy ($U$\%, $S$\%, $H$\%) on the four datasets, where Inf. means inference-based methods and Gen. means generation-based methods. Note that PS V2.0 was released recently in \url{https://drive.google.com/file/d/1p9gtkuHCCCyjkyezSarCw-1siCSXUykH/view} to fix a test-data leaking bug. This can have large impacts on the performance of existing methods, such as GDAN~\cite{huang2019generative}. Therefore all our evaluations were conducted on PS V2.0. $\dagger$ indicates that the results are taken from the PS V2.0 report~\cite{xian2020zero}, and we reproduced the results on all other methods using the official code. For our method GCM-CF, we used AREN~\cite{xie2019attentive} for Conventional ZSL on CUB, and otherwise used TF-VAEGAN~\cite{narayan2020latent} for supervised classification and Conventional ZSL.}
\label{tab:3}
\vspace{-0.3cm}
\end{table*}

\subsection{Evaluation Metrics and Settings}
\label{sec:4.2}
\noindent\textbf{ZSL Evaluation}. It was conducted in the Generalized ZSL setting. We used two metrics: 1) \textbf{ZSL Accuracy}. It consists of 3 numbers $(U,S,H)$, where $U$/$S$ is the per-class top-1 accuracy of unseen-/seen-classes test samples, and $H$ is the harmonic mean of $U,S$, given by $H=2\times S \times U / (S+U)$. 2) \textbf{CVb}. To measure the balance between unseen/seen classification, we propose to use the Coefficient of Variation of the seen and unseen binary classification accuracy, denoted as CVb. Let $S_b$ and $U_b$ be the binary accuracy on seen- and unseen-classes, respectively. CVb is given by $\sqrt{0.5(S_b-\mu)^2 + 0.5(U_b-\mu)^2} / \mu$, where $\mu=(S_b+U_b)/2$. Note that the variation between $S$ and $U$ of ZSL Accuracy is not a good measure of balance, as they are affected by the number of seen- and unseen-classes, which can be quite different (see SUN in Table~\ref{tab:1}). 3) \textbf{AUSUC}. 
We draw the Seen-Unseen accuracy Curve (SUC) by plotting a series of $S$ against $U$ of ZSL Accuracy, where the series is obtained by adjusting a calibration factor $\omega$ that is subtracted from the classifier logits on the seen-classes. Then we use the Area Under SUC (AUSUC) for evaluation. Compared to a single ZSL Accuracy, SUC and the area provide a more detailed view of the capability of an algorithm to balance the unseen-seen decision boundary~\cite{chao2016empirical,chen2018zero}.

\noindent\textbf{OSR Evaluation}. We used the following metrics: 
1) Macro-averaged \textbf{F1 scores} over seen-classes and ``unknown'' (for all unseen-classes), which shows how well a method can recognize seen classes while rejecting unseen-classes samples; 2) \textbf{Openness-F1 Plot}. We also studied the response of F1 scores under varying \textit{openness} given by $1-\sqrt{2N / (N+M)}$, where $N$ and $M$ are number of seen- and unseen-classes, respectively. Compared to a single F1 score where the openness is fixed, this plot shows the robustness of an OSR classifier to the open environment with an unknown number of unseen-classes.

\noindent\textbf{Implementation Details}. For ZSL, we implemented our GCM based on the network architecture in TF-VAEGAN~\cite{narayan2020latent}. Following common protocol~\cite{xian2018feature,li2019leveraging}, we used the ResNet-101~\cite{he2016deep} features for $X$ and attributes provided in~\cite{xian2018zero} for $\mathcal{Y}_S,\mathcal{Y}_U$. For OSR, our GCM was implemented using the networks in CGDL~\cite{sun2020conditional} and $X$ represents actual images. Other details are in Appendix.

\subsection{Results on ZSL}
\label{sec:4.3}


\begin{table}[h!]
\centering
\captionsetup{font=footnotesize,labelfont=footnotesize,skip=5pt}
\scalebox{0.8}{
\begin{tabular}{p{3.5cm} p{1cm}<{\centering}p{1cm}<{\centering}p{1cm}<{\centering}p{1cm}<{\centering}}
\hline\hline
\multicolumn{1}{c}{\large{Method}} & CUB & AWA2 & SUN & aPY\\
\hline
GDAN~\cite{WelinderEtal2010} & 15.1 & 25.1 & 27.3 & 14.4 \\
CADA-VAE~\cite{xiao2010sun} & 6.7 & 6.0 & 7.3 & 11.2 \\
LisGAN~\cite{xian2018zero} & 4.9 & 7.0 & 4.6 & 4.0 \\
TF-VAEGAN~\cite{farhadi2009describing} & 8.0 & 9.4 & 10.2 & 5.8\\
\textbf{GCM-CF (Ours)} & \cellcolor{mygray}\textbf{1.5} & \cellcolor{mygray}\textbf{2.1} & \cellcolor{mygray} \textbf{1.0} & \cellcolor{mygray} \textbf{2.3}\\
\hline \hline
\end{tabular}}
\caption{CVb (\%) of generative models on all datasets.}
\label{tab:4}
\vspace{-0.3cm}
\end{table}
\begin{table}[t]
\centering
\captionsetup{font=footnotesize,labelfont=footnotesize,skip=5pt}
\scalebox{0.8}{
\begin{tabular}{p{2.7cm} p{0.01cm}<{\centering}p{0.5cm}<{\centering}p{0.5cm}<{\centering}p{0.5cm}<{\centering}p{0.01cm}<{\centering}p{0.5cm}<{\centering}p{0.5cm}<{\centering}p{0.5cm}<{\centering}}
\hline\hline
\multicolumn{1}{c}{\multirow{2}{*}{\diagbox[height=6.45ex, width=9.5em]{Stage 2}{Stage 1}}} &     & \multicolumn{3}{c}{\textbf{TF-VAEGAN~\cite{narayan2020latent}}} & &              \multicolumn{3}{c}{\textbf{GCM-CF (Ours)}} \\ \cmidrule(lr){3-5}\cmidrule(lr){7-9}
 &  & $U$   & $S$    & $H$      &    & $U$ & $S$   & $H$\\ \hline

    RelationNet~\cite{sung2018learning} & & 49.3 & 81.2 & 61.3 & & 55.8  & 75.0 & \textbf{64.0}\\

    CADA-VAE~\cite{schonfeld2019generalized}  &  & 49.5 & 81.1 & 61.5  &  & 57.6 & 75.0 & \textbf{65.2} \\
    LisGAN~\cite{li2019leveraging} & & 48.8 & 80.4 & 60.7 & & 56.1  & 74.3 & \textbf{63.9}\\
    
    TF-VAEGAN~\cite{narayan2020latent} & & 52.8 & 83.2 & 64.6 &  & 60.4 & 75.1 & \textbf{67.0}\\
\hline \hline
\end{tabular}}
\caption{Comparison of the two-stage inference performance on AWA2~\cite{xian2018zero} using TF-VAEGAN~\cite{narayan2020latent} and our GCM-CF as the stage-one binary classifier. The results on other datasets are shown in Appendix, where using GCM-CF as stage-one binary classifier improves all the methods.}
\label{tab:5}
\vspace{-0.3cm}
\end{table}

\begin{figure}
    \centering
    \captionsetup{font=footnotesize,labelfont=footnotesize,skip=5pt}
    \includegraphics[width=0.95\linewidth]{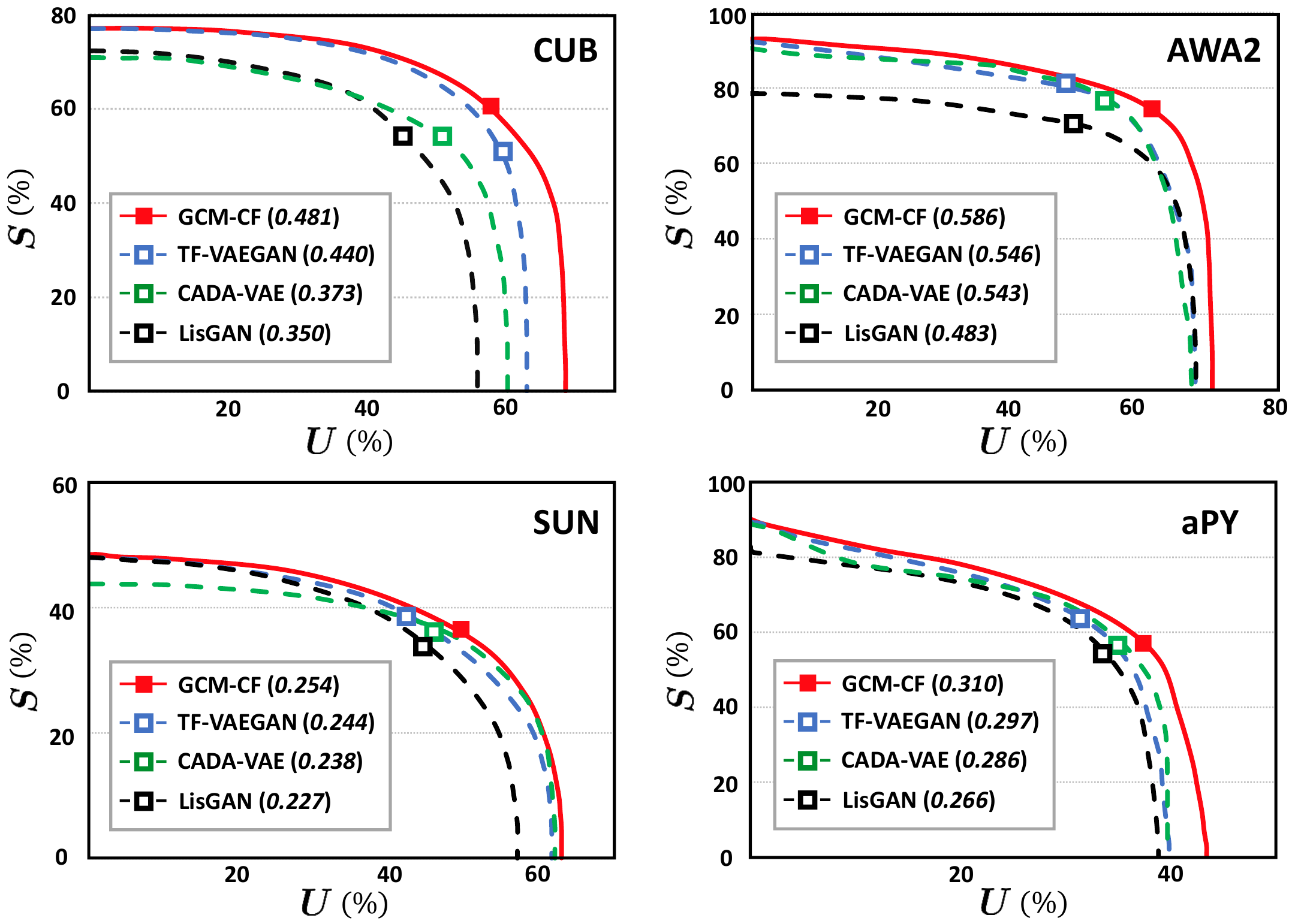}
    \caption{The Seen-Unseen accuracy Curve (SUC) together with the Area Under SUC (AUSUC) on the four datasets. The rectangle denotes the ZSL Accuracy when calibration factor $\omega$=0.}
    \vspace*{-4.1mm}
    \label{fig:6}
\end{figure}

\noindent\textbf{Mitigate the Imbalance}. As shown in Table~\ref{tab:3}, our counterfactual approach, denoted as GCM-CF, achieves a \textbf{more balanced ZSL Accuracy} and significantly improves the existing state-of-the-art (SOTA) by 2.2\% to 4.3\%, with a much higher score on $U$. For example, compared to LisGAN on aPY, our method gains 3.9\% on $U$ while sacrificing only 0.1\% on $S$. To further show that GCM-CF mitigates the unseen/seen imbalance, we diagnosed the binary classification accuracy using CVb in Table~\ref{tab:4}. Note that existing methods have very large CVb, which means that there is a large difference between seen and unseen classification accuracy and reveals the imbalance problem. Our method has the \textbf{lowest CVb}. This shows that our approach indeed achieves a more balanced binary decision boundary between seen and unseen. However, one may argue that the imbalance problem can be solved by simply adjusting the calibration factor $\omega$. Therefore, we plot the SUC using varying $\omega$ and measured the AUSUC on all datasets. The result is shown in Figure~\ref{fig:6}, where GCM-CF outperforms other methods in every inch and achieves \textbf{the best AUSUC}. This shows that GCM-CF fundamentally improves the unseen/seen classification beyond the reach of simple calibration. Overall, the balanced and much improved ZSL Accuracy, lower CVb for binary classification, and higher accuracy on varying calibrations demonstrate that our method mitigates the unseen/seen imbalance in ZSL. This indicates that our GCM generates faithful counterfactuals (see Figure~\ref{fig:2c}) and supports the effectiveness of our counterfactual-faithful training in disentangling $Z$ and $Y$.

\noindent\textbf{Stage-One Binary Classifier}. Our GCM-CF can serve as a stage-one binary unseen/seen classifier and plug into \emph{all} ZSL methods for subsequent supervised learning on seen and conventional ZSL on unseen. We performed experiments on 4 representative methods---inference based: RelationNet~\cite{sung2018learning}, generation with VAE: CADA-VAE~\cite{schonfeld2019generalized}, with GAN: LisGAN~\cite{li2019leveraging} and with VAE-GAN~\cite{larsen2016autoencoding}: TF-VAEGAN~\cite{narayan2020latent}. The ZSL Accuracy on AWA2 is shown in Table~\ref{tab:5}. By comparing with Table~\ref{tab:3}, we observe that GCM-CF can significantly improve all of them on $H$. For comparison, we used the current SOTA ZSL method TF-VAEGAN as a binary unseen/seen classifier and performed the same experiments. The results on the ZSL Accuracy show that our GCM-CF significantly outperforms TF-VAEGAN. This demonstrates that compared to the class-agnostic $Z$ in existing methods, the use of sample attribute $Z$ that is disentangled from class attribute $Y$ in our GCM-CF is highly effective in improving ZSL performance. Overall, our method as a robust binary classifier can serve as a new strong baseline to evaluate ZSL methods using two-stage inference.

\begin{figure*}
    \centering
    \captionsetup{font=footnotesize,labelfont=footnotesize,skip=5pt}
    \includegraphics[width=\textwidth]{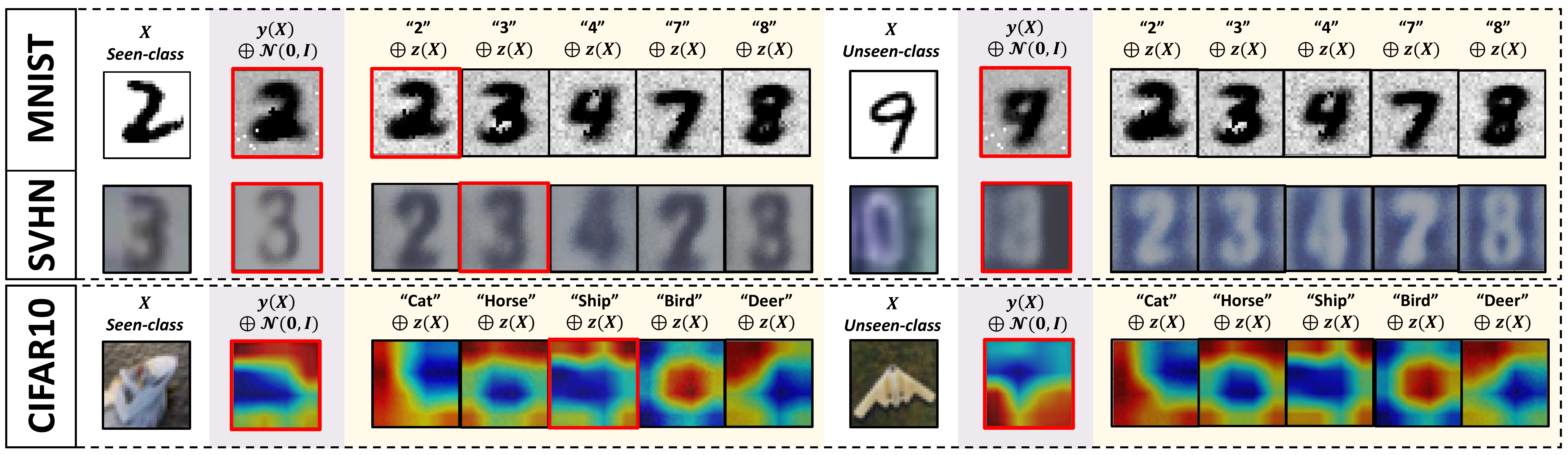}
    \caption{Comparison of the reconstructed images using CGDL~\cite{sun2020conditional} ($y(X) \oplus \mathcal{N}(0,I)$) and the counterfactual images generated from our GCM-CF ($y\oplus z(X)$) on the seen- and unseen-class samples. The red box on a generated image denotes that it is similar to the original. For CIFAR10, due to the conflict between visual perception and discriminative training~\cite{ilyas2019adversarial}, we first train a classifier on seen-classes images and then use CAM~\cite{zhou2016learning} to show the sensible yet discrimiantive features: although the pixel-level generation is not sensible, the pixel-level distance remains discriminative as shown in Table~\ref{tab:6}.}
    \label{fig:7}
    \vspace*{-1mm}
\end{figure*}

\subsection{Results on OSR}
\label{sec:4.4}

\begin{figure}
    \vspace*{-4.5mm}
    \captionsetup{font=footnotesize,labelfont=footnotesize,skip=5pt}
    \begin{minipage}[c]{0.63\linewidth}
        \includegraphics[width=\linewidth]{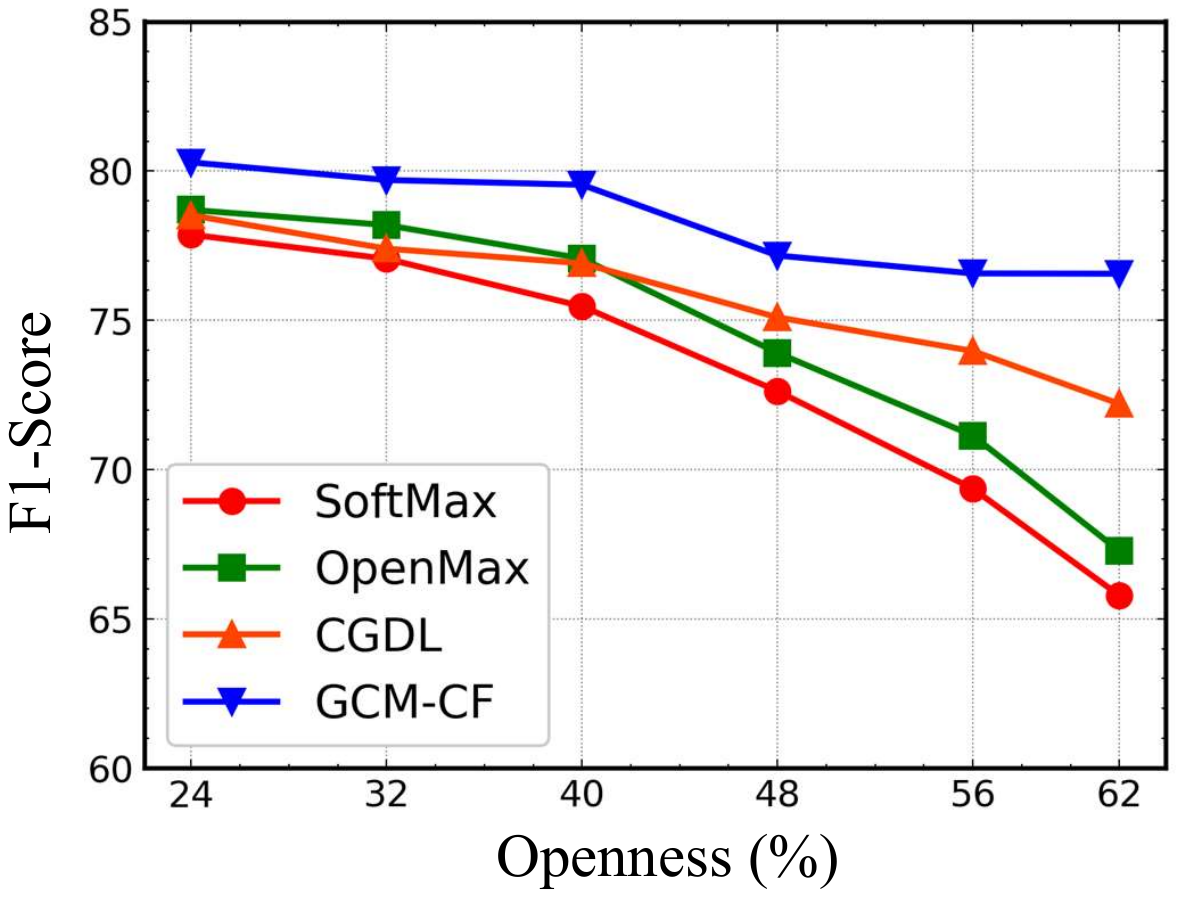}
    \end{minipage}\hfill
    \begin{minipage}[c]{0.37\linewidth}
        \captionof{figure}{Openness-F1 Plot where 4 non-animal classes from CIFAR10~\cite{krizhevsky2009learning} were used as seen-classes and various classes were drawn from CIFAR100~\cite{krizhevsky2009learning} as unseen-classes.}
        \label{fig:openness}
    \end{minipage}
    \vspace*{-4.8mm}
\end{figure}

\noindent\textbf{Strong Open-Set Classifier}. In Table~\ref{tab:6}, our GCM-CF achieves SOTA F1 scores in all datasets. We discovered that the common evaluation setting~\cite{neal2018open,yoshihashi2019classification} of averaging F1 scores over 5 random splits can result in a large variance in the F1 score. Therefore, we additionally show the results on all 5 splits in the Appendix, where GCM-CF outperforms other methods in every split. This strongly supports that GCM-CF improves F1 scores in general, \ie, not only on some particular splits. So, where does the improvement come from? To rule out the possibility that F1 improves due to a higher classification accuracy on the seen-classes, we measure the Closed-Set Accuracy of a standard CNN trained in a supervised fashion and that of CGDL, which was used in GCM-CF for 2nd stage supervised classification. The results are shown in the Appendix where the performances are similar. Therefore, the increased F1 score indeed comes from improved unseen/seen binary classification. Furthermore, a strong open-set classifier should stay robust regardless of the number of unseen-classes during evaluation. Therefore, we evaluate OSR classifiers with the Openness-F1 Plot. As shown in Figure~\ref{fig:openness}, our GCM-CF achieves the highest F1 score on all openness settings, and our performance is especially competitive in the challenging environment with large openness. Overall, these results clearly demonstrate the effectiveness of our GCM-CF on seen-unseen classification in OSR.

\begin{table}[t]
\centering
\vspace*{-1.5mm}
\captionsetup{font=footnotesize,labelfont=footnotesize,skip=5pt}
\scalebox{0.8}{
\begin{tabular}{p{2.55cm} p{1.0cm}<{\centering}p{1.0cm}<{\centering}p{1.2cm}<{\centering}p{1.0cm}<{\centering}p{1.0cm}<{\centering}}
\hline\hline
\multicolumn{1}{c}{\large{Method}} & MNIST & SVHN & CIFAR10 & C+10 & C+50 \\
\hline
Softmax & 76.82 & 76.16 & 70.39 & 77.82 & 65.96 \\
OpenMax~\cite{bendale2016towards} & 85.93 & 77.95 & 71.38 & 78.68 & 67.68 \\
CGDL~\cite{sun2020conditional} & 88.95 & 76.31 & 71.03 & 77.92 & 70.96 \\
\textbf{GCM-CF (Ours)} & \cellcolor{mygray}\textbf{91.37} & \cellcolor{mygray}\textbf{79.25} & \cellcolor{mygray}\textbf{72.63} & \cellcolor{mygray}\textbf{79.38} & \cellcolor{mygray}\textbf{74.60} \\
\hline \hline
\end{tabular}}
\caption{Comparison of the F1 score averaged over 5 random splits in OSR. We used the official code on CGDL~\cite{sun2020conditional} and implemented Softmax and OpenMax~\cite{bendale2016towards} for evaluation. For GCM-CF, after binary classification, we used CGDL for supervised classification on the seen-classes.}
\label{tab:6}
\vspace*{-4mm}
\end{table}

\noindent\textbf{Qualitative Results}. Figure~\ref{fig:7} shows OSR evaluation using existing reconstruction-based approach and our counterfactual approach. Notice the reconstructed image on the unseen-class sample (\eg, ``0, 9'') can be similar to the original image. This makes it difficult to tell apart unseen and seen by thresholding reconstruction error. By using our GCM-CF to generate counterfactuals on each seen-class, the generated images are counterfactual-faithful and indeed look like seen-classes. Note that this holds on CIFAR10, where given seen- or unseen-class samples, the CAMs of the generated counterfactuals using the same seen-class attribute look similar (more results in the Appendix). Therefore, by thresholding the dissimilarity between the test sample and its counterfactuals, we can easily distinguish unseen from seen. These results show the effectiveness of our binary inference strategy. Interestingly, we also observe that the generated samples from GCM indeed capture sample attributes better (\eg, text color, background color on SVHN), which validates that we can learn a disentangled $Z$ using the proposed counterfactual-faithful training.

\vspace{-3mm}
\section{Conclusions}
\label{sec:5}

We presented a novel counterfactual framework for Zero-Shot Learning (ZSL) and Open-Set Recognition (OSR) to provide a theoretical ground for balancing and improving the seen/unseen classification imbalance. Specifically, we proposed a Generative Causal Model to generate faithful counterfactuals, which allows us to use the Consistency Rule for balanced binary seen/unseen classification. Extensive results in ZSL and OSR show that our method indeed improves the balance and hence achieves the state-of-the-art performance. As future direction, we will seek new definitions on disentanglement~\cite{suter2019robustly} and devise practical implementations to achieve improved disentanglement~\cite{besserve2020theory}.
\section{Acknowledgements}

The authors would like to thank all the anonymous reviewers for their constructive comments and suggestions. This research is partly supported by the Alibaba-NTU Singapore Joint Research Institute, Nanyang Technological University (NTU), Singapore; the Singapore Ministry of Education (MOE) Academic Research Fund (AcRF) Tier 1 and Tier 2 grant; and Alibaba Innovative Research (AIR) programme. We also want to thank Alibaba City Brain Group for the donations of GPUs.

{\small
\bibliographystyle{ieee_fullname}
\bibliography{references}
}

\clearpage
\renewcommand{\thesection}{A.\arabic{section}}
\renewcommand*{\thesubsection}{A.\arabic{section}.\arabic{subsection}}
\renewcommand{\thetable}{A\arabic{table}}
\renewcommand{\thefigure}{A\arabic{figure}}
\setcounter{section}{0}
\setcounter{figure}{0}
\setcounter{table}{0}

\noindent\textbf{\Large Appendix}
\vspace{0.1in}

This appendix is organized as follows:

\begin{itemize}
    \item Section~\ref{sec:a1} proves the theorem in Section 3.4 as a corollary of~\cite{besserve2020counterfactual};
    \item Section~\ref{sec:a2} gives the implementation details of our model in ZSL (Section~\ref{sec:a2.1}) and OSR (Section~\ref{sec:a2.2});
    \item Section~\ref{sec:a3} includes additional experimental results; Specifically, Section~\ref{sec:a3.1} shows additional results on two-stage inference and a comparison between entangled and disentangled model in ZSL Accuracy. Section~\ref{sec:a3.2} shows additional results for OSR, with the details of the 5 splits that we used and the F1 score for each split, closed-set accuracy, a comparison between entangled and disentangled model and more qualitative results.
\end{itemize}

\section{Proof of the Theorem in 3.4}
\label{sec:a1}

Let $V=(Z,Y)$ and $V$ takes values from the space $\mathcal{V}$. For class attribute $Y$ that is a $K$-dimensional real vector, denote $\mathcal{E}=\{e_1,\ldots,e_K\}$ as the subset of dimensions spanned by the class attribute $Y$, and $\bar{\mathcal{E}}$ as those by the sample attribute $Z$. Therefore, $\mathcal{V}^\mathcal{E}$ and $\mathcal{V}^{\bar{\mathcal{E}}}$ represent the space of class attribute $Y$ and sample attribute $Z$, respectively. We use $g:\mathcal{V}\to \mathcal{X}$ to denote the endogenous mapping to the feature space $\mathcal{X}$. Note that this $g$ corresponds to sampling from $P_\theta(X|Z,Y)$ in our GCM. We will introduce the concept of embedded GCM to facilitate the proof.

\noindent\textbf{Definition} (Embedded GCM). \emph{We say that a GCM is embedded if $g:\mathcal{V}\to \mathcal{X}$ is a continuous injective function with continuous inversion $g^{-1}$.}

Using the results in~\cite{armstrong2013basic}, if $V$ is compact (\ie, bounded), the GCM is embedded if and only if $g$ is injective. Though we implement our GCM using VAE, whose latent space is not compact, it is shown that restricting a VAE to a product of compact intervals that covers most of the probability mass (using KL-divergence in the objective) will result in an embedded GCM that approximates the original one for most samples~\cite{besserve2020counterfactual}. Moreover, $P_\theta(X|Z,Y)$ are implemented using deterministic mappings in our model (see Section~\ref{sec:a2.1},~\ref{sec:a2.2}), which are indeed injective as shown in~\cite{puthawala2020globally}. Therefore, without loss of generality, our GCM can be considered as embedded. We will give the formal definition of intrinsic disentanglement, which can be used to show that group disentanglement of $Z$ and $Y$ leads to faithfulness.

\noindent\textbf{Definition} (Intrinsic Disentanglement). \emph{In a GCM, the endomorphism $T:\mathcal{X}\to\mathcal{X}$ is intrinsically disentangled with respect to the subset $\mathcal{E}$ of endogenous variables, if there exists a transformation $T'$ affecting only variables indexed by $\mathcal{E}$, such that for any $\mathbf{v}\in \mathcal{V}$,}
\begin{equation}
    T(g(\mathbf{v}))=g(T'(\mathbf{v})).
\end{equation}

Now we will first establish the equivalence between intrinsic disentanglement and faithfulness using the theorem below.

\noindent\textbf{Theorem} (Intrinsic Disentanglement and Faithfulness). \emph{The counterfactual mapping $X_{\mathbf{y}}[z(X)]$ is faithful if and only if it is intrinsically disentangled with respect to the subset $\mathcal{E}$.}

To prove the above theorem, one conditional is trivial: if
a transformation is intrinsically disentangled, it is by definition an endomorphism of $\mathcal{X}$ so the counterfactual mapping must be faithful. For the second conditional, let us assume a faithful counterfactual mapping $X_{\mathbf{y}}[z(X)]$. Based on the three steps of computing counterfactuals and the embedding property discussed earlier, the counterfactual mapping can be decomposed as:
\begin{equation}
    X_{\mathbf{y}}[z(X)]=g \circ T' \circ g^{-1}(X),
\end{equation}
where $\circ$ denotes function composition, $T'$ is the counterfactual transformation of $V$ as defined in Section 3.2, where $Z$ is kept as $Z=z(X)$ and $Y$ is set as $\mathbf{y}$. Now for any $\mathbf{v}\in \mathcal{V}$, the quantity $X_{\mathbf{y}}[z(g(\mathbf{v}))]$ can be similarly decomposed as:
\begin{equation}
    X_{\mathbf{y}}[z(g(\mathbf{v}))] = g \circ T' \circ g^{-1} \circ g(\mathbf{v}) = g \circ T'(\mathbf{v}).
\end{equation}
Since $T'$ is a transformation that only affects variables in $\mathcal{E}$ (\ie, Y), we show that faithful counterfactual transformation $X_{\mathbf{y}}[z(X)]$ is intrinsically disentangled with respect to $\mathcal{E}$.

In our work, we define group disentanglement of $Z$ and $Y$ as intrinsic disentanglement with respect to the set of variables in $Y$. We used the sufficient condition, \ie, learning a GCM where $Y$ and $Z$ are group disentangled, such that when we fix $Z$ and only change $Y$, the resulting generation lies in $\mathcal{X}$ according to the theorem that we just proved. 

\section{Implementation Details}
\label{sec:a2}

\subsection{ZSL}
\label{sec:a2.1}
Our GCM implementation is based on the generative models in TF-VAEGAN~\cite{narayan2020latent}. Besides $P_\theta(X|Z,Y)$ and $Q_\phi(Z|X)$ that is common in a VAE-GAN~\cite{larsen2016autoencoding}, it additionally implements $Q_\psi(Y|X)$ and a feedback module, which takes the intermediate layer output from the network in $Q_\psi(Y|X)$ as input and generate a feedback to assist the generation process $P_\theta(X|Z,Y)$. The rest of the section will detail the network architecture for each component, followed by additional training and inference details supplementary to Section 3.4 and 3.3, respectively.

\noindent\textbf{Sample Attribute $Z$}. The dimension of sample attribute $Z$ is set to be the same as that of the class attribute for each dataset. For example, in CUB~\cite{WelinderEtal2010}, the dimension of $Z$ is set as 312. $P(Z)$ is defined as $\mathcal{N}(\mathbf{0},\mathbf{I})$ for all datasets.

\noindent\textbf{Decoder $P_\theta(X|Z,Y)$}. The module that implements this conditional distribution is commonly known as the decoder in literature~\cite{schonfeld2019generalized,narayan2020latent}. We implemented $P_\theta(X|Z,Y)$ with Deep Gaussian Family $\mathcal{N}(\mu_D(Z,Y),\mathbf{I})$ with its variance fixed and mean generated from a network $\mu_D$. $\mu_D$ was implemented with a Multi-Layer Perceptron (MLP) with two layers and LeakyReLU activation (alpha=0.2) in between. The input to the MLP is the concatenated $Z$ and $Y$. The hidden layer size is set as 4,096. The MLP outputs a real vector of size 2,048 (same as that of $X$) and the output goes through a Sigmoid activation to produce the mean of $P_\theta(X|Z,Y)$.

\noindent\textbf{Encoder $Q_\phi(Z|X)$}. For convenience, we refer to $Q_\phi(Z|X)$ as the encoder. We implemented $Q_\phi(Z|X)$ with $\mathcal{N}(\mu_E(X),\sigma^2_E(X))$, where $\mu_E(X),\sigma^2_E(X)$ are neural networks that share identical architecture. Specifically, they are 3-layer MLP with LeakyReLU activation (alpha=0.2) that takes $X$ as input and outputs a vector with the same dimension as $Z$. The first hidden layer size is set as 4,096 and the second hidden layer size is set as two times of the dimension of $Z$. Note that in the original TF-VAEGAN implementation, the encoder additionally conditions on $Y$. We argue that this may cause the encoded $Z$ to contain information about $Y$, undermining the disentanglement. Hence we make the encoder conditioned only on $X$.

\noindent\textbf{Regressor $Q_\psi(Y|X)$}. It is a 2-layer MLP with LeakyReLU activation (alpha=0.2). The hidden layer size is set as 4,096.

\noindent\textbf{Discriminator $D(X,Y)$}. It is a 2-layer MLP with LeakyReLU activation (alpha=0.2). It takes the concatenated $X$ and $Y$ as input and outputs a single real value. The hidden layer size is set as 4,096.

\noindent\textbf{Feedback Module}.  It is a 2-layer MLP with LeakyReLU activation (alpha=0.2). The hidden layer output of the regressor is sent to the feedback module as input. This module generates a real vector with 4,096 dimensions, which is added to the hidden layer output of $\mu_D$ as feedback signal.

\noindent\textbf{Training}. The networks are trained in an iterative fashion. First, all networks except the decoder are optimized. Then the discriminator is frozen and all other networks are optimized. We followed the optimization settings in TF-VAEGAN. Specifically, the Adam~\cite{kingma2014adam} optimizer is used with learning rate set as $1e^{-4}$ in CUB~\cite{WelinderEtal2010}, $1e^{-5}$ in AWA2~\cite{xian2018zero}, $1e^{-3}$ in SUN~\cite{xiao2010sun} and $1e^{-5}$ in aPY~\cite{farhadi2009describing}. For the hyperparameters in our counterfactual-faithful training, we used $\beta=6.0, \nu=1.0$ for CUB, $\beta=6.0, \nu=1.0$ for AWA2, $\beta=4.0, \nu=1.0$ for SUN and $\beta=6.0, \nu=0$ for aPY. On CUB, AWA2, and SUN, we additionally used annealing on the KL divergence loss, where $\beta$ is initially set as 0 and linearly increased to the set value over 40 epochs. The parameter $\rho$ is set to $1.0$ in ZSL task.

\noindent\textbf{Inference}. In ZSL, we train a linear classifier with one fully-connected layer using the Adam optimizer. On CUB, the classifier was trained for 15 epochs with learning rate as $1e^{-3}$. On AWA2, it was trained for 3 epochs with learning rate as $1e^{-3}$. On SUN, it was trained for 6 epochs with learning rate as $5e^{-4}$. On aPY, it was trained for 3 epochs with learning rate as $1e^{-3}$. After training, the classifier was used for inference following the decision rule introduced in Section 3.3.

\subsection{OSR}

Our proposed GCM-CF is implemented based on the architecture of CGDL~\cite{sun2020conditional}. Notice that the original CGDL doesn't distinguish sample attribute $Z$ and class attribute $Y$ explicitly. To keep consistent with the ZSL model and follow the common VAE-GAN architecture, here we revise the encoder to model $Z$ and $Y$ respectively.

\noindent\textbf{Encoder $Q_\phi(Z|X)$}.
Given an actual image $X=\bm{x}$, we follow \cite{sun2020conditional} to implement $Q_\phi(Z|X)$ with the probabilistic ladder architecture to extract the high-level abstract latent features $\bm{z}$.
In detail, the $l$-th layer in the ladder encoder is expressed as:
\begin{equation*}
\begin{aligned}
&\bm{x}_l=\text{Conv}(\bm{x}_{l-1})\\
&\bm{h}_l=\text{Flatten}(\bm{x}_l)\\
&\bm{\mu}_l=\text{Linear}(\bm{h}_l)\\
&\bm{\sigma}^{2}_l=\text{Softplus}(\text{Linear}(\bm{h}_l))
\end{aligned}
\end{equation*} 
where \verb Conv  is a convolutional layer followed by a batch-norm layer and a PReLU layer, \verb Flatten  is a linear layer to flatten 2-dimensional data into 1-dimension, \verb Linear  is a single linear layer and \verb Softplus  applies $\log(1+\text{exp}(\cdotp))$ non-linearity to each component of its argument vector. The latent representation $Z=\bm{z}$ can be obtained as:
\begin{equation}
\begin{aligned}
    \bm{\mu_z}, \bm{\sigma^2_z} &= \text{LadderEncoder} (\bm{x}) , \\
    \bm{z} &= \mu_z + \sigma_z \odot \mathcal{N}(0,\mathbf{I}) ,
\end{aligned}
\end{equation}
where $\odot$ is the element-wise product.

\noindent\textbf{Decoder $P_\theta(X|Z,Y)$}.
Given the latent sample attribute $Z=\bm{z}$ and the class attribute $Y=\bm{y}$, the $l$-th layer of the ladder decoder is expressed as follows:
\begin{equation*}
\begin{aligned}
&\bm{\tilde c}_{l+1}=\text{Unflatten}(\bm{t}_{l+1})\\
&\bm{\tilde x}_{l+1}=\text{ConvT}(\bm{\tilde c}_{l+1})\\
&\bm{\tilde h}_{l+1}=\text{Flatten}(\bm{\tilde x}_{l+1})\\
&\bm{\tilde \mu}_l=\text{Linear}(\bm{\tilde h}_{l+1})\\
&\bm{\tilde \sigma}^{2}_l=\text{Softplus}(\text{Linear}(\bm{\tilde h}_{l+1}))\\
&\bm{t}_l=\bm{\tilde \mu}_l+\bm{\tilde \sigma}^{2}_l\odot \bm{\epsilon}
\end{aligned}
\end{equation*} 
where \verb ConvT  is a transposed convolutional layer and \verb Unflatten  is a linear layer to convert 1-dimensional data into 2-dimension. Note that the input $\bm{t}$ of the top decoder layer is the concatenation of $\bm{z}$ and $\bm{y}$. Overall, the reconstructed image $\bm{\tilde x}$ can be represented as:
\begin{equation}
    \bm{\tilde x} = \text{LadderDecoder} (\bm{z}, \bm{y}).
\end{equation}
For more details please refer to \cite{sun2020conditional}.

\noindent\textbf{Known Classifier}.
The known classifier is a Softmax Layer taking the one-hot embedding $\bm{y}$ as the input and produces the probability distribution over the known classes.

\noindent\textbf{Training}.
The network is trained in an end-to-end fashion. We directly follow the optimization settings and hyperparameters of ladder architecture in CGDL~\cite{sun2020conditional}. For the hyperparameters in our counterfactual-faithful training, we used $\beta=1.0, \nu=10.0$ for MNIST, $\beta=1.0, \nu=2.0$ for SVHN, $\beta=6.0, \nu=1.0$ for CIFAR10, $\beta=1.0, \nu=20.0$ for CIFARAdd10, $\beta=1.0, \nu=1.0$ for CIFARAdd50. Note that the $\rho$ is set to $0$ in OSR task.

\noindent\textbf{Inference}.
When training is completed, we follow \cite{sun2020conditional} to use the reconstruction errors and a multivariate Gaussian model to judge the unseen samples. The threshold $\tau_l$ is set to 0.9 for MNIST, 0.6 for SVHN, 0.9 for CIFAR10, 0.8 for CIFARAdd10 and 0.5 for CIFARAdd50. More details about the multivariate Gaussian model please refer to \cite{sun2020conditional}.

\label{sec:a2.2}

\section{Additional Results}
\label{sec:a3}

\subsection{ZSL}
\label{sec:a3.1}

\begin{figure}
    \centering
    \footnotesize
    \includegraphics[width=.48\textwidth]{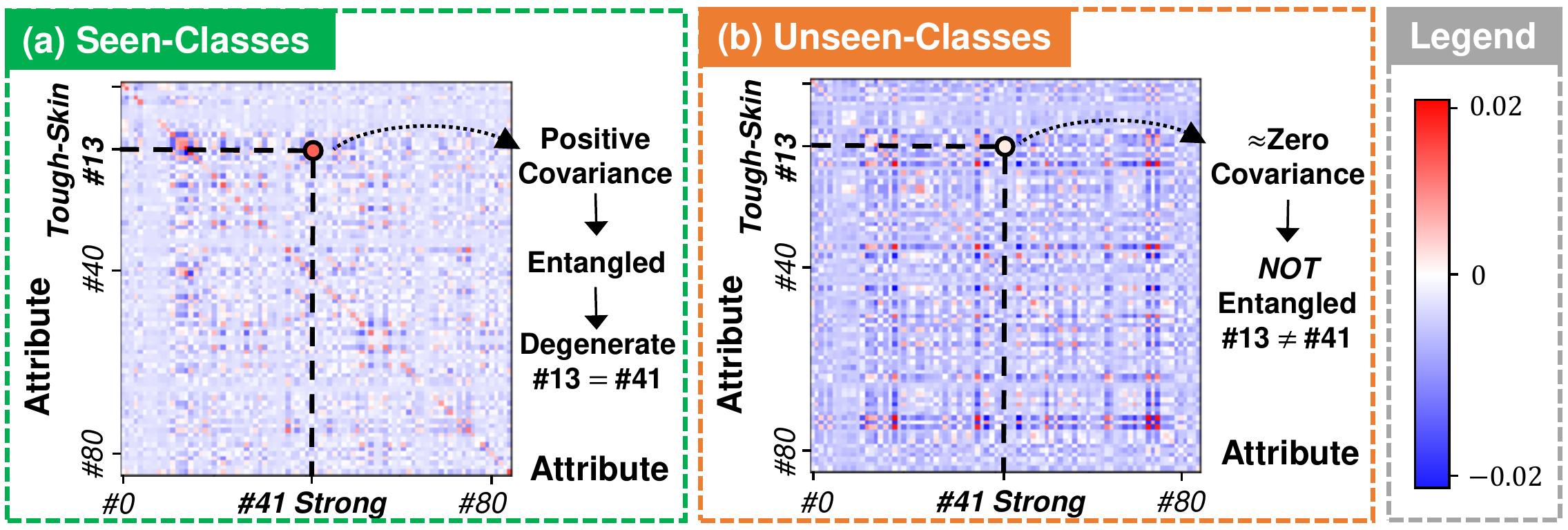}
    \caption{Visualization of the covariance matrix in AWA2 dataset of the (a) seen-classes and (b) unseen-classes attributes.}
    \vspace{-5mm}
    \label{fig:y_entangled}
\end{figure}

\noindent\textbf{Entanglement of $Y$}. Two class attributes are entangled when they always or never appear together, and they effectively degenerate to one feature. If two attributes are entangled among seen-classes but not unseen, the degenerated feature learnt from seen cannot tell unseen-classes apart. Entanglement can be identified by finding pairs of attributes with large absolute covariance. For example, Figure~\ref{fig:y_entangled} shows a ``strong'' animal usually has ``tough skin'' among seen-classes. Yet the co-appearance no longer holds for unseen-classes as shown in Figure~\ref{fig:y_entangled}.

\noindent\textbf{Stage-One Binary Classifier}. We extend the results in Table 5 by showing comparison of the two-stage inference performance on CUB~\cite{WelinderEtal2010}, SUN~\cite{xiao2010sun} and aPY~\cite{farhadi2009describing} dataset. Our GCM-CF improves all of them and outperforms the current SOTA ZSL method TF-VAEGAN as a binary unseen/seen classifier.

\begin{table}[t]
\centering
\captionsetup{font=footnotesize,labelfont=footnotesize,skip=5pt}
\scalebox{0.8}{
\begin{tabular}{p{2.7cm} p{0.01cm}<{\centering}p{0.5cm}<{\centering}p{0.5cm}<{\centering}p{0.5cm}<{\centering}p{0.01cm}<{\centering}p{0.5cm}<{\centering}p{0.5cm}<{\centering}p{0.5cm}<{\centering}}
\hline\hline
\multicolumn{1}{c}{Dataset} & \multicolumn{8}{c}{CUB~\cite{WelinderEtal2010}}\\
\hline
\multicolumn{1}{c}{\multirow{2}{*}{\diagbox[height=6.45ex, width=9.5em]{Stage 2}{Stage 1}}} &     & \multicolumn{3}{c}{\textbf{TF-VAEGAN~\cite{narayan2020latent}}} & &              \multicolumn{3}{c}{\textbf{GCM-CF (Ours)}} \\ \cmidrule(lr){3-5}\cmidrule(lr){7-9}
 &  & $U$   & $S$    & $H$      &    & $U$ & $S$   & $H$\\ \hline

    RelationNet~\cite{sung2018learning} & & 40.5 & 65.3 & 50.0 & & 47.7  & 57.6 & \textbf{52.2}\\

    CADA-VAE~\cite{schonfeld2019generalized}  &  & 43.2 & 63.4 & 51.4  &  & 51.4 & 57.6 & \textbf{54.3} \\
    LisGAN~\cite{li2019leveraging} & & 41.1 & 66.0 & 50.7 & & 47.9  & 58.1 & \textbf{52.5}\\
    
    TF-VAEGAN~\cite{narayan2020latent} & & 50.8 & 64.0 & 56.6 &  & 55.4 & 60.0 & \textbf{57.6}\\
\hline \hline

\multicolumn{1}{c}{Dataset} & \multicolumn{8}{c}{SUN~\cite{xiao2010sun}}\\
\hline
\multicolumn{1}{c}{\multirow{2}{*}{\diagbox[height=6.45ex, width=9.5em]{Stage 2}{Stage 1}}} &     & \multicolumn{3}{c}{\textbf{TF-VAEGAN~\cite{narayan2020latent}}} & &              \multicolumn{3}{c}{\textbf{GCM-CF (Ours)}} \\ \cmidrule(lr){3-5}\cmidrule(lr){7-9}
 &  & $U$   & $S$    & $H$      &    & $U$ & $S$   & $H$\\ \hline

    RelationNet~\cite{sung2018learning} & & 30.8 & 23.0 & 26.3 & & 37.2  & 21.9 & \textbf{27.6}\\

    CADA-VAE~\cite{schonfeld2019generalized}  &  & 37.6 & 39.3 & 38.4 &  & 44.6 & 37.6 & \textbf{40.8} \\
    LisGAN~\cite{li2019leveraging} & & 36.3 & 41.7 & 38.8 & & 43.0  & 38.9 & \textbf{40.8}\\
    
    TF-VAEGAN~\cite{narayan2020latent} & & 41.7 & 39.9 & 40.8 &  & 47.9 & 37.8 & \textbf{42.2}\\
\hline \hline

\multicolumn{1}{c}{Dataset} & \multicolumn{8}{c}{aPY~\cite{farhadi2009describing}}\\
\hline
\multicolumn{1}{c}{\multirow{2}{*}{\diagbox[height=6.45ex, width=9.5em]{Stage 2}{Stage 1}}} &     & \multicolumn{3}{c}{\textbf{TF-VAEGAN~\cite{narayan2020latent}}} & &              \multicolumn{3}{c}{\textbf{GCM-CF (Ours)}} \\ \cmidrule(lr){3-5}\cmidrule(lr){7-9}
 &  & $U$   & $S$    & $H$      &    & $U$ & $S$   & $H$\\ \hline

    RelationNet~\cite{sung2018learning} & & 31.5 & 63.3 & 42.1 & & 34.6  & 56.6 & \textbf{43.0}\\

    CADA-VAE~\cite{schonfeld2019generalized}  &  & 31.1 & 64.8 & 41.9  &  & 35.0 & 57.2 & \textbf{43.5} \\
    LisGAN~\cite{li2019leveraging} & & 31.2 & 64.6 & 42.0 & & 34.7  & 57.3 & \textbf{43.2}\\
    
    TF-VAEGAN~\cite{narayan2020latent} & & 33.1 & 64.2 & 43.7 &  & 37.1 & 56.8 & \textbf{44.9}\\
\hline \hline
\end{tabular}}
\caption{Supplementary to Table 5. Comparison of the two-stage inference performance on CUB~\cite{WelinderEtal2010}, SUN~\cite{xiao2010sun} and aPY~\cite{farhadi2009describing} using TF-VAEGAN~\cite{narayan2020latent} and our GCM-CF as the stage-one binary classifier.}
\label{tab:5ext}
\vspace{-0.3cm}
\end{table}

\noindent\textbf{Effect of Disentanglement}. To show that the quality of disentangling $Z$ and $Y$ is the key bottleneck, we compared an entangled model (without counterfactual-faithful training) and the disentangled model on ZSL Accuracy and the results are shown in Table~\ref{tab:disentangled}. Notice that the entangled model has a much lower $S$. This is because the training is conducted on the seen-classes and the encoded $Z$ is entangled with seen-classes attributes. Therefore the generated counterfactuals are biased towards the seen-classes, \ie, the green counterfactual samples in Figure 2c are closer to true samples from seen-classes, pushing the classifier boundary towards seen-classes and increasing the recall of the unseen-class by sacrificing
that of the seen.

\begin{table}[t]
\centering
\captionsetup{font=footnotesize,labelfont=footnotesize,skip=5pt}
\scalebox{0.8}{
\begin{tabular}{p{2cm} p{0.01cm}<{\centering}p{0.5cm}<{\centering}p{0.5cm}<{\centering}p{0.5cm}<{\centering}p{0.01cm}<{\centering}p{0.5cm}<{\centering}p{0.5cm}<{\centering}p{0.5cm}<{\centering}}
\hline\hline
\multicolumn{1}{c}{\multirow{2}{*}{Dataset}} &     & \multicolumn{3}{c}{\textbf{Entangled}} & &              \multicolumn{3}{c}{\textbf{Disentangled}} \\ \cmidrule(lr){3-5}\cmidrule(lr){7-9}
 &  & $U$   & $S$    & $H$      &    & $U$ & $S$   & $H$\\ \hline

    CUB~\cite{WelinderEtal2010} & & 71.6 & 15.8 & 25.9 & & 61.0  & 59.7 & \textbf{60.3}\\

    AWA2~\cite{xian2018zero}  &  & 70.5 & 27.9 & 40.0  &  & 60.4 & 75.1 & \textbf{67.0} \\
    SUN~\cite{xiao2010sun} & & 62.9 & 17.5 & 27.4 & & 47.9  & 37.8 & \textbf{42.2}\\
    
    aPY~\cite{farhadi2009describing} & & 42.4 & 14.4 & 21.5 &  & 37.1 & 56.8 & \textbf{44.9}\\
\hline \hline
\end{tabular}}
\caption{Comparison of ZSL Accuracy using an entangled model without using the proposed counterfactual-faithful training and the disentangled model with the proposed training.}
\label{tab:disentangled}
\end{table}

\subsection{OSR}
\label{sec:a3.2}

\noindent\textbf{5 Splits Results}. In our main paper we have argued that the common evaluation setting of averaging F1 scores over 5 random splits can result in a large variance in the F1 score. Here we additionally report the split details in Table~\ref{tab:split_detail} and the results on all splits in Tabel~\ref{tab:a6}. Note that since the official code of CGDL~\cite{sun2020conditional} is not complete, we implemented the code of dataloader, computing F1 score and set part of parameters. The 5 seeds (\ie, 5 splits) are randomly chosen without any selection.

\begin{table}[th]
\centering
\captionsetup{font=footnotesize,labelfont=footnotesize,skip=5pt}
\scalebox{0.75}{
\begin{tabular}{lm{2cm}<{\centering}m{5.5cm}<{\centering}}
\hline\hline
\multicolumn{1}{c}{Split}    & \multicolumn{2}{c}{1 (seed: 777)} \\
\hline
\multicolumn{1}{c}{Dataset}    & Seen                    & Unseen                        \\ \hline
MNIST      & 3,7,8,2,4,6    & 0,1,5,9                              \\ \hline
SVHN       & 3,7,8,2,4,6   & 0,1,5,9                                                                      \\ \hline
CIFAR10    & 3,7,8,2,4,6             & 0,1,5,9                                                          \\ \hline
CIFARAdd10 & 0,1,8,9                 & 27, 46, 98, 38, 72, 31, 36, 66, 3, 97                             \\ \hline
CIFARAdd50 & 0,1,8,9 & 27, 46, 98, 38, 72, 31, 36, 66, 3, 97, 75, 67, 42, 32, 14, 93, 6, 88, 11, 1, 44, 35, 73, 19, 18, 78, 15, 4, 50, 65, 64, 55, 30, 80, 26, 2, 7, 34, 79, 43, 74, 29, 45, 91, 37, 99, 95, 63, 24, 21 \\ 
\hline\hline

\multicolumn{1}{c}{Split}    & \multicolumn{2}{c}{2 (seed: 1234)} \\
\hline
\multicolumn{1}{c}{Dataset}    & Seen                    & Unseen                        \\ \hline
MNIST      & 7,1,0,9,4,6    & 2,3,5,8                              \\ \hline
SVHN       & 7,1,0,9,4,6   & 2,3,5,8                                                                      \\ \hline
CIFAR10    & 7,1,0,9,4,6             & 2,3,5,8                                                          \\ \hline
CIFARAdd10 & 0,1,8,9                 & 98, 46, 14, 1, 7, 73, 3, 79, 93, 11 
\\ \hline
CIFARAdd50 & 0,1,8,9   & 98, 46, 14, 1, 7, 73, 3, 79, 93, 11, 37, 29, 2, 74, 91, 77, 55, 50, 18, 80, 63, 67, 4, 45, 95, 30, 75, 97, 88, 36, 31, 27, 65, 32, 43, 72, 6, 26, 15, 42, 19, 34, 38, 66, 35, 21, 24, 99, 78, 44 \\ 
\hline\hline

\multicolumn{1}{c}{Split}    & \multicolumn{2}{c}{3 (seed: 2731)} \\
\hline
\multicolumn{1}{c}{Dataset}    & Seen                    & Unseen                        \\ \hline
MNIST      & 8,1,6,7,2,4    & 0,3,5,9                              \\ \hline
SVHN       & 8,1,6,7,2,4   & 0,3,5,9                                                                      \\ \hline
CIFAR10    & 8,1,6,7,2,4             & 0,3,5,9                                                          \\ \hline
CIFARAdd10 & 0,1,8,9                 & 79, 98, 67, 7, 77, 42, 36, 65, 26, 64                             \\ \hline
CIFARAdd50 & 0,1,8,9 & 79, 98, 67, 7, 77, 42, 36, 65, 26, 64, 66, 73, 75, 3, 32, 14, 35, 6, 24, 21, 55, 34, 30, 43, 93, 38, 19, 99, 72, 97, 78, 18, 31, 63, 29, 74, 91, 4, 27, 46, 2, 88, 45, 15, 11, 1, 95, 50, 80, 44 \\ 
\hline\hline

\multicolumn{1}{c}{Split}    & \multicolumn{2}{c}{4 (seed: 3925)} \\
\hline
\multicolumn{1}{c}{Dataset}    & Seen                    & Unseen                        \\ \hline
MNIST      & 7,3,8,4,6,1    & 0,2,5,9                              \\ \hline
SVHN       & 7,3,8,4,6,1   & 0,2,5,9                                                                      \\ \hline
CIFAR10    & 7,3,8,4,6,1             & 0,2,5,9                                                          \\ \hline
CIFARAdd10 & 0,1,8,9                 & 46, 77, 29, 24, 65, 66, 79, 21, 1, 95                             \\ \hline
CIFARAdd50 & 0,1,8,9 & 46, 77, 29, 24, 65, 66, 79, 21, 1, 95, 36, 88, 27, 99, 67, 19, 75, 42, 2, 73, 32, 98, 72, 97, 78, 11, 14, 74, 50, 37, 26, 64, 44, 30, 31, 18, 38, 4, 35, 80, 45, 63, 93, 34, 3, 43, 6, 55, 91, 15 \\ 
\hline\hline

\multicolumn{1}{c}{Split}    & \multicolumn{2}{c}{5 (seed: 5432)} \\
\hline
\multicolumn{1}{c}{Dataset}    & Seen                    & Unseen                        \\ \hline
MNIST      & 2,8,7,3,5,1    & 0,4,6,9                              \\ \hline
SVHN       & 2,8,7,3,5,1   & 0,4,6,9                                                                      \\ \hline
CIFAR10    & 2,8,7,3,5,1             & 0,4,6,9                                                          \\ \hline
CIFARAdd10 & 0,1,8,9                 & 21, 95, 64, 55, 50, 24, 93, 75, 27, 36                             \\ \hline
CIFARAdd50 & 0,1,8,9 & 21, 95, 64, 55, 50, 24, 93, 75, 27, 36, 73, 63, 19, 98, 46, 1, 15, 72, 42, 78, 77, 29, 74, 30, 14, 38, 80, 45, 4, 26, 31, 11, 97, 7, 66, 65, 99, 34, 6, 18, 44, 3, 35, 88, 43, 91, 32, 67, 37, 79 \\ 
\hline\hline

\end{tabular}}
\caption{The detailed label splits of 5 random seeds}
\label{tab:split_detail}
\end{table}
\begin{table}[h]
\centering
\captionsetup{font=footnotesize,labelfont=footnotesize,skip=5pt}
\scalebox{0.8}{
\begin{tabular}{p{2.55cm} p{1.0cm}<{\centering}p{1.0cm}<{\centering}p{1.2cm}<{\centering}p{1.0cm}<{\centering}p{1.0cm}<{\centering}}
\hline\hline
\multicolumn{1}{c}{Split} & \multicolumn{5}{c}{1} \\
\hline
\multicolumn{1}{c}{Method} & MNIST & SVHN & CIFAR10 & C+10 & C+50 \\
\hline
Softmax & 76.26 & 73.06 & 69.81 & 77.87 & 65.78 \\
OpenMax~\cite{bendale2016towards} & 83.34 & 75.34 & 71.49 & 78.70 & 67.27 \\
CGDL~\cite{sun2020conditional} & 91.79 & 77.42 & 70.02 & 78.52 & 72.7 \\
\textbf{GCM-CF (Ours)} & \textbf{94.21} & \textbf{79.23} & \textbf{73.03} & \textbf{80.29} & \textbf{74.70} \\
\hline \hline

\multicolumn{1}{c}{Split} & \multicolumn{5}{c}{2} \\
\hline
\multicolumn{1}{c}{Method} & MNIST & SVHN & CIFAR10 & C+10 & C+50 \\
\hline
Softmax & 77.06 & 75.03 & 73.02 & 77.82 & 65.87 \\
OpenMax~\cite{bendale2016towards} & 86.97 & 77.27 & 73.74 & 79.02 & 67.56 \\
CGDL~\cite{sun2020conditional} & 86.76 & 73.63 & 73.15 & 76.46 & 70.79 \\
\textbf{GCM-CF (Ours)} & \textbf{91.82} & \textbf{80.28} & \textbf{75.71} & \textbf{79.67} & \textbf{74.79} \\
\hline \hline

\multicolumn{1}{c}{Split} & \multicolumn{5}{c}{3} \\
\hline
\multicolumn{1}{c}{Method} & MNIST & SVHN & CIFAR10 & C+10 & C+50 \\
\hline
Softmax & 77.44 & 78.67 & 70.79 & 77.61 & 66.21 \\
OpenMax~\cite{bendale2016towards} & 83.39 & 80.00 & 72.01 & 78.38 & 67.83 \\
CGDL~\cite{sun2020conditional} & 92.36 & 77.59 & 74.77 & 77.92 & 71.93 \\
\textbf{GCM-CF (Ours)} & \textbf{93.86} & \textbf{80.51} & \textbf{75.38} & \textbf{79.40} & \textbf{76.56} \\
\hline \hline

\multicolumn{1}{c}{Split} & \multicolumn{5}{c}{4} \\
\hline
\multicolumn{1}{c}{Method} & MNIST & SVHN & CIFAR10 & C+10 & C+50 \\
\hline
Softmax & 76.03 & 75.47 & 70.25 & 77.67 & 66.01 \\
OpenMax~\cite{bendale2016towards} & 87.06 & 76.80 & 70.76 & 78.64 & 68.21 \\
CGDL~\cite{sun2020conditional} & 90.34 & 73.53 & 70.30 & 78.27 & 71.69 \\
\textbf{GCM-CF (Ours)} & \textbf{91.34} & \textbf{80.76} & \textbf{71.12} & \textbf{78.74} & \textbf{74.53} \\
\hline \hline

\multicolumn{1}{c}{Split} & \multicolumn{5}{c}{5} \\
\hline
\multicolumn{1}{c}{Method} & MNIST & SVHN & CIFAR10 & C+10 & C+50 \\
\hline
Softmax & 77.33 & 78.30 & 68.12 & 78.13 & 65.97 \\
OpenMax~\cite{bendale2016towards} & 89.88 & 80.33 & 68.90 & 78.70 & 67.55 \\
CGDL~\cite{sun2020conditional} & 83.51 & 79.41 & 66.89 & 78.00 & 68.18 \\
\textbf{GCM-CF (Ours)} & \textbf{92.45} & \textbf{80.33} & \textbf{69.52} & \textbf{78.79} & \textbf{72.40} \\
\hline \hline
\end{tabular}}
\caption{Comparison of the F1 score averaged over 5 random splits in OSR. Note that since the official code of CGDL~\cite{sun2020conditional} is not complete, we implemented the code of dataloader, F1 score and set part of parameters. Moreover, we also implemented Softmax and OpenMax~\cite{bendale2016towards} for evaluation. For GCM-CF, after binary classification, we used CGDL for supervised classification on the seen-classes.}
\label{tab:a6}
\end{table}

\noindent\textbf{Closed Set Results}.
Closed-Set Accuracy is the standard supervised classification accuracy on the seen-classes with open set detection disabled. As shown in Table~\ref{tab:closeset}, the network were trained without any large degradation in closed set accuracy from the plain CNN.

\begin{table}[t]
\centering
\captionsetup{font=footnotesize,labelfont=footnotesize,skip=5pt}
\scalebox{0.83}{
\begin{tabular}{p{2.0cm} p{1.1cm}<{\centering}p{1.1cm}<{\centering}p{1.1cm}<{\centering}p{1.1cm}<{\centering}p{1.1cm}<{\centering}}
\hline\hline
\large{Method} & MNIST & SVHN & CIFAR10 & C+10 & C+50 \\
\hline
Plain CNN & 0.995 & 0.965 & 0.917 & 0.941 & 0.940 \\
CGDL~\cite{sun2020conditional} & 0.996 & 0.962 & 0.913 & 0.934 & 0.935 \\
\hline \hline
\end{tabular}}
\caption{Comparison of the Closed-Set accuracy in OSR.}
\label{tab:closeset}
\end{table}

\noindent\textbf{Effect of disentanglement}.
To further demonstrate the effectiveness of disentangling $Z$ and $Y$ in OSR, we also compared an entangled model (without the counterfactual-faithful training) and the disentangled model on F1 scores. The results are shown in Table~\ref{tab:osr_disentangle}. Similar to the ZSL, we can also see the F1 scores of entangled model are much lower than those of disentangled model. The constructed green counterfactual samples are still closer to the unseen sample though given the seen attributes without disentanglement, which demonstrate the necessity of the proposed disentangle loss.
\begin{table}[t]
\centering
\captionsetup{font=footnotesize,labelfont=footnotesize,skip=5pt}
\scalebox{0.83}{
\begin{tabular}{p{2.0cm} p{1.1cm}<{\centering}p{1.1cm}<{\centering}p{1.1cm}<{\centering}p{1.1cm}<{\centering}p{1.1cm}<{\centering}}
\hline\hline
\large{Model} & MNIST & SVHN & CIFAR10 & C+10 & C+50 \\
\hline
Entangled & 91.37 & 62.57 & 67.03 & 73.81 & 69.18 \\
Disentangled & \textbf{94.21} & \textbf{79.23} & \textbf{73.03} & \textbf{80.29} & \textbf{74.70} \\
\hline \hline
\end{tabular}}
\caption{Comparison of the F1 scores using entangeled and disentangled model in OSR.}
\label{tab:osr_disentangle}
\end{table}

\begin{figure*}[ht]
    \centering
    \captionsetup{font=footnotesize,labelfont=footnotesize,skip=5pt}
    \includegraphics[width=\textwidth]{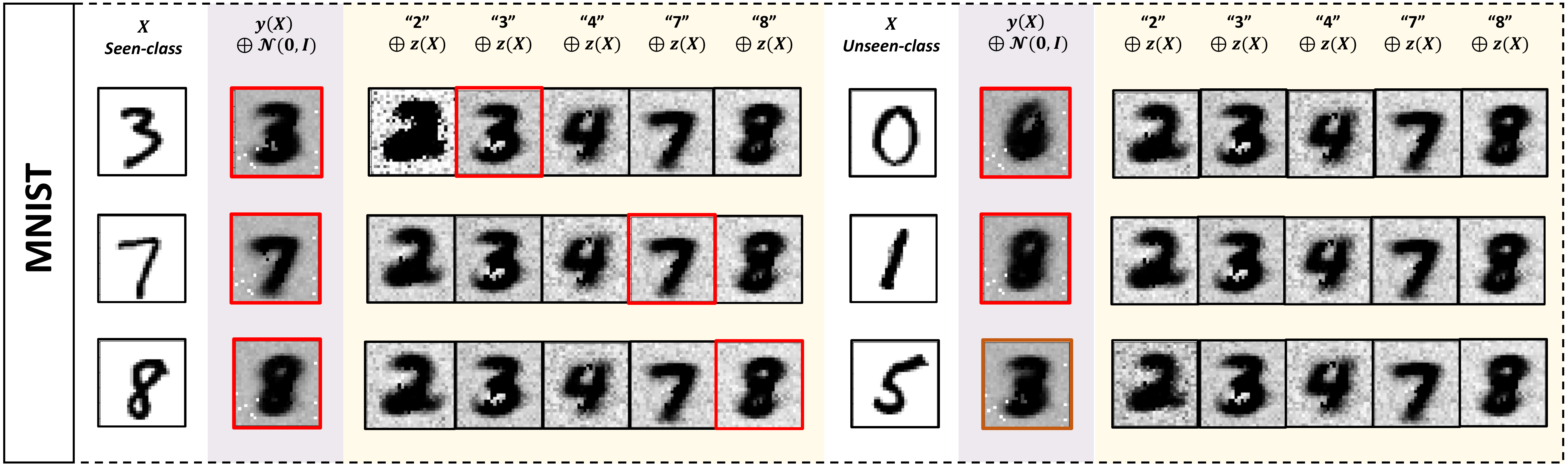}
    \caption{The additional qualitative results of the reconstructed images using CGDL~\cite{sun2020conditional} ($y(X) \oplus \mathcal{N}(0,I)$) and the counterfactual images generated from our GCM-CF ($y\oplus z(X)$) on MNIST dataset. The red box on a generated image denotes that it is similar to the original, while the brown box represents the failure generation.}
    \label{fig:qua_mnist}
    \vspace*{-1mm}
\end{figure*}
\begin{figure*}[ht]
    \centering
    \captionsetup{font=footnotesize,labelfont=footnotesize,skip=5pt}
    \includegraphics[width=\textwidth]{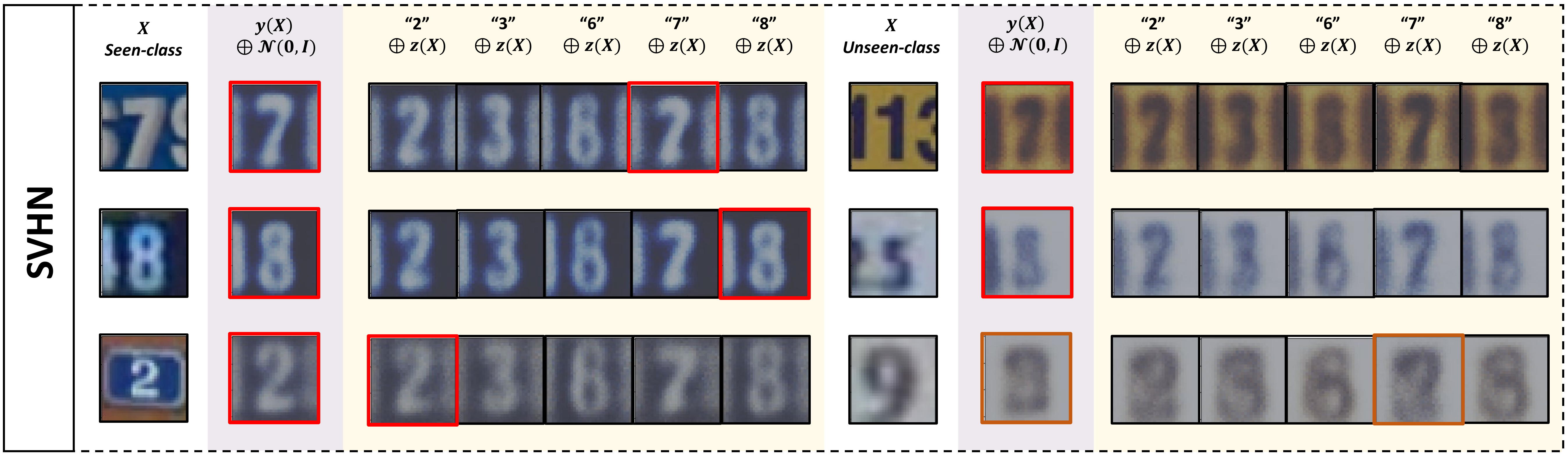}
    \caption{The additional qualitative results of the reconstructed images using CGDL~\cite{sun2020conditional} ($y(X) \oplus \mathcal{N}(0,I)$) and the counterfactual images generated from our GCM-CF ($y\oplus z(X)$) on SVHN dataset. The red box on a generated image denotes that it is similar to the original, while the brown box represents the failure generation.}
    \label{fig:qua_svhn}
    \vspace*{-1mm}
\end{figure*}

\noindent\textbf{More Qualitative Results}.
Figure~\ref{fig:qua_mnist} and \ref{fig:qua_svhn} show the additional qualitative results comparing existing reconstruction-based approach with our proposed counterfactual approach on MNIST and SVHN dataset. For the \textit{Seen-class} (\ie, the left part), both the baseline model and our GCM-CF can reconstruct samples with low reconstruction error, which means both of them can handle well given the seen-class images.
When coming to the unseen-class (\ie, the right part), the baseline method would still generate similar samples (red box), with a much lower reconstruction error comparing to the counterfactual samples produced by GCM-CF, resulting in a failure rejection to the unknown outlier.
The brown box represents the failure reconstructions for the baseline model (\ie, generated sample is also dissimilar with the original input image) given some unseen-class samples. Note that this is reasonable since the model haven't seen the unseen-class samples during training, which also corresponds to Figure~4b in the main paper. In this case, our counterfactual model can still make better generation (\eg, ``3'' in the last row of Figure~\ref{fig:qua_mnist}).

\begin{figure*}
    \centering
    \captionsetup{font=footnotesize,labelfont=footnotesize,skip=5pt}
    \includegraphics[width=\textwidth]{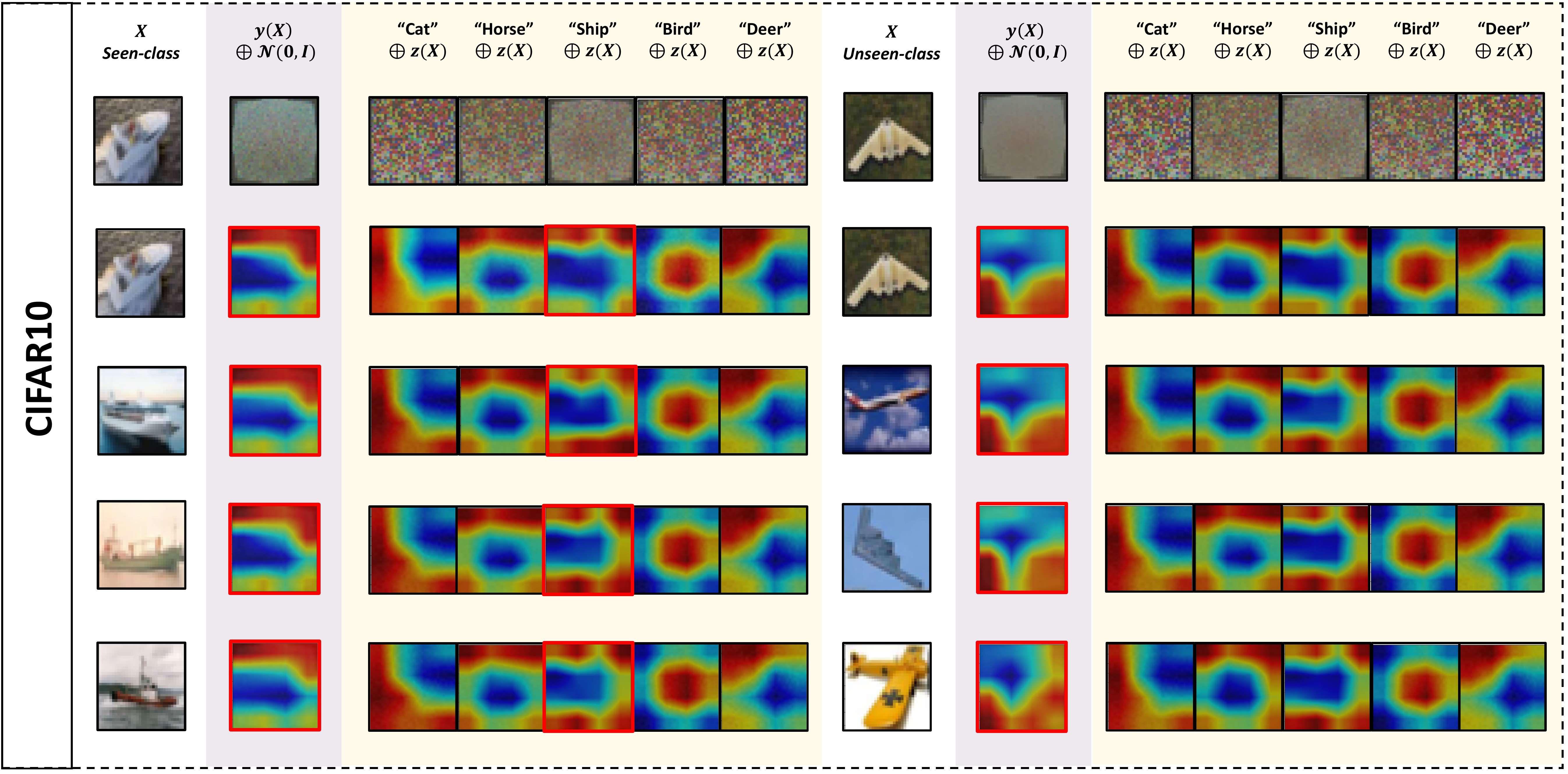}
    \caption{The additional qualitative results of the reconstructed images using CGDL~\cite{sun2020conditional} ($y(X) \oplus \mathcal{N}(0,I)$) and the counterfactual images generated from our GCM-CF ($y\oplus z(X)$) on CIFAR10 dataset. The red box on a generated image denotes that it is similar to the original.}
    \label{fig:qua_cifar}
    \vspace*{-1mm}
\end{figure*}

For the CIFAR10 dataset, as dicussed in the main paper, we cannot generate realistic images due to the conflict between visual perception and discriminative training. Therefore, we apply a pretrained image classifier to generate CAM to reveal the sensible yet discriminating features. Here we show the additional examples in Figure~\ref{fig:qua_cifar}. The first row is the direct image reconstruction results generated by baseline and proposed model. 
The disordered appearance explains that the pixel-level generation is not sensible. However, when utilizing the tool of CAM, something magical happened.
The insensible pixel-level generation becomes discriminative in the view of the pretrained classifier. The generation samples of our proposed GCM-CF reveal different heat maps given different counterfactual class conditions. Among them the heat map of ``ship'' condition is quite similar to that of the reconstruction of the baseline model (red box), showing the consistency of the CAM heat map.
Moreover, for different samples in the same class (\ie, the different rows), the class-specific CAM heat maps keep stable with only minor changes. It further demonstrates that the CAM heat map can be considered a kind of substitution of the original pixel images to reveal the discrimitative feature.

\end{document}